\def\year{2022}\relax

\documentclass[letterpaper]{article} 
\usepackage{aaai22} 
\usepackage{times}  
\usepackage{helvet}  
\usepackage{courier}  
\usepackage[hyphens]{url}  
\usepackage{graphicx} 
\usepackage{natbib}  
\usepackage{caption} 
\DeclareCaptionStyle{ruled}{labelfont=normalfont,labelsep=colon,strut=off} 
\usepackage[ruled, vlined]{algorithm2e}

\usepackage{hyperref}
\usepackage{placeins}
\usepackage{times}
\usepackage{latexsym}
\usepackage{graphicx}
\usepackage{subfigure}
\usepackage{booktabs} 
\usepackage{amsmath}
\usepackage{amssymb}
\usepackage{amsthm}
\usepackage{pifont}
\usepackage{xcolor}

\nocopyright

\newcommand{\txb}{\ensuremath{\tilde{\mathbf{x}}}}
\newcommand{\xb}{\ensuremath{\mathbf{x}}}

\newcommand{\yb}{\ensuremath{\mathbf{y}}}

\newcommand{\cmark}{\ding{51}}%
\newcommand{\xmark}{\ding{55}}%

\SetKwInput{KwGiven}{Define}
\SetKwInput{KwGiven}{Given}
\SetKwInput{KwParams}{Parameters}

\title{Symbolic Brittleness in Sequence Models: \\
on Systematic Generalization in Symbolic Mathematics}

\author {
    Sean Welleck,\textsuperscript{\rm 1 \rm 2}
    Peter West,\textsuperscript{\rm 1}
    Jize Cao,\textsuperscript{\rm 1}
    Yejin Choi\textsuperscript{\rm 1 \rm 2}
}
\affiliations {
    \textsuperscript{\rm 1} Paul G. Allen School of Computer Science \& Engineering, University of Washington\\
    \textsuperscript{\rm 2} Allen Institute for Artificial Intelligence\\
    wellecks@uw.edu
}

\begin{document}

\maketitle

\begin{abstract}
Neural sequence models trained with maximum likelihood estimation have led to breakthroughs in many tasks, where success is defined by the gap between training and test performance.
However, their ability to achieve stronger forms of generalization remains unclear.
We consider the problem of symbolic mathematical integration, as it requires generalizing systematically beyond the test set.
We develop a methodology for evaluating generalization that takes advantage of the problem domain's structure and access to a verifier.
Despite promising in-distribution performance of sequence-to-sequence models in this domain, we demonstrate challenges in achieving robustness, compositionality, and out-of-distribution generalization, through both carefully constructed manual test suites and a genetic algorithm that automatically finds large collections of failures in a controllable manner. 
Our investigation highlights the difficulty of generalizing well with the predominant modeling and learning approach, and the importance of evaluating beyond the test set, across different aspects of generalization.\footnote{Code available at:\\ \url{https://github.com/wellecks/symbolic_generalization}}
\end{abstract}

\section{Introduction}

Despite their success, recent studies reveal undesirable properties of conventional neural sequence models, such as assigning high-probabilities to unrealistic sequences \citep{holtzman2020The,welleck2020consistency}, susceptibility to adversarial attacks \citep{wallace2019universal}, and limited generalization on symbolic tasks \citep{saxton2019analysing,nogueira2021InvestigatingTL}, even with very large models and datasets \citep{henighan2020ScalingLF}.
Despite these drawbacks, \citet{lample2019deep} recently demonstrated that a standard sequence-to-sequence model, which we call a \textit{neural sequence integrator}, performs surprisingly well at \textit{symbolic integration}, solving problems that are beyond traditional symbolic solvers and achieving near perfect test accuracy.

\begin{table}[t!]
\footnotesize
\setlength{\tabcolsep}{6pt}
\begin{center}
\begin{tabular}{lllr}
\toprule
\textbf{Input} & \textbf{Integral} & \textbf{Prediction} & \\
\midrule
$30\cos(39x)$ &  $\frac{10}{13}\sin(39x)$ & $\frac{10}{13}\sin(39x)$ & \cmark\\[3pt]
$17\cos(83x)$ & $\frac{17}{83}\sin(83x)$ & ${\color{red}{\frac{1}{17}}\sin(83x)}$ &\xmark\\[3pt]
$34\cos(77x)$ & $\frac{34}{77}\sin(77x)$ & ${\color{red}\sin(77x)}$ & \xmark\\[3pt]
\hline 
\rule{0pt}{1.2\normalbaselineskip}$x^{209}$& $\frac{1}{210}x^{210}$  & $\frac{1}{210}x^{210}$ & \cmark\\[3pt]
$x^{764}$& $\frac{1}{765}x^{765}$ & $\frac{1}{765}x^{765}$ & \cmark\\[3pt]
$x^{209}+x^{764}$& $\frac{1}{210}x^{210}+\frac{1}{765}x^{765}$& ${\color{red}{\frac{1}{205}x^{205}}}$ &\xmark\\[3pt]
\hline 
\rule{0pt}{1.2\normalbaselineskip} $-241$ & $-241x$ & ${\color{red}{-239x}} - 14400$ & \xmark\\[3pt]
$123^x$ & $\frac{123^x}{\log(123)}$ & ${\color{red}{\frac{123^x}{1 + \log(123)}}}$ & \xmark\\[3pt]
$4^x + x^{465} + 1$ & $\frac{x^{466}}{466} + x + \frac{4^x}{\log (4)}$ & $\frac{x^{466}}{466} + x + {\color{red}{e^x}}$ & \xmark \\
\bottomrule
\end{tabular}
\caption{\label{tbl:front-examples} 
Despite its impressive ability to integrate equations that are out of reach for traditional symbolic solvers, the neural sequence integrator shows deficiences in \textbf{robustness} (top) and \textbf{compositionality} (middle), and fails on \textbf{adversarial} problems discovered by SAGGA (bottom).
}
\end{center}
\end{table}

Recent studies suggest that achieving \textit{strong and systematic generalization} is difficult with vanilla sequence-to-sequence methods, as they latch onto regularities in the training data, learning dataset-specific solutions that do not generalize beyond the training distribution (e.g. \citet{agrawal2016analyzing,lake2018still,bahdanau2019,hupkes2020}).
Symbolic integration -- finding the integral of a mathematical function -- specifically requires these forms of generalization, as it involves an underlying structure that extends beyond this fixed training distribution. 
For instance, the rule $\int k = kx+C$ applies to all constants $k$, and the sum rule $\int f_1+\int f_2=\int (f_1+f_2)$ means that integrating two functions correctly should imply integrating their sum correctly.
Symbolic integration also offers a structured problem domain and a verifier for evaluating whether a proposed solution is correct, making it an effective setting for evaluating generalization.
As the neural sequence integrator relies on a common recipe-- a large-scale transformer trained to maximize the likelihood of a training set of input-output sequences -- it is especially interesting to study whether it generalizes systematically.


\begin{figure}
\begin{center}
\footnotesize
\includegraphics[width=0.99\linewidth]{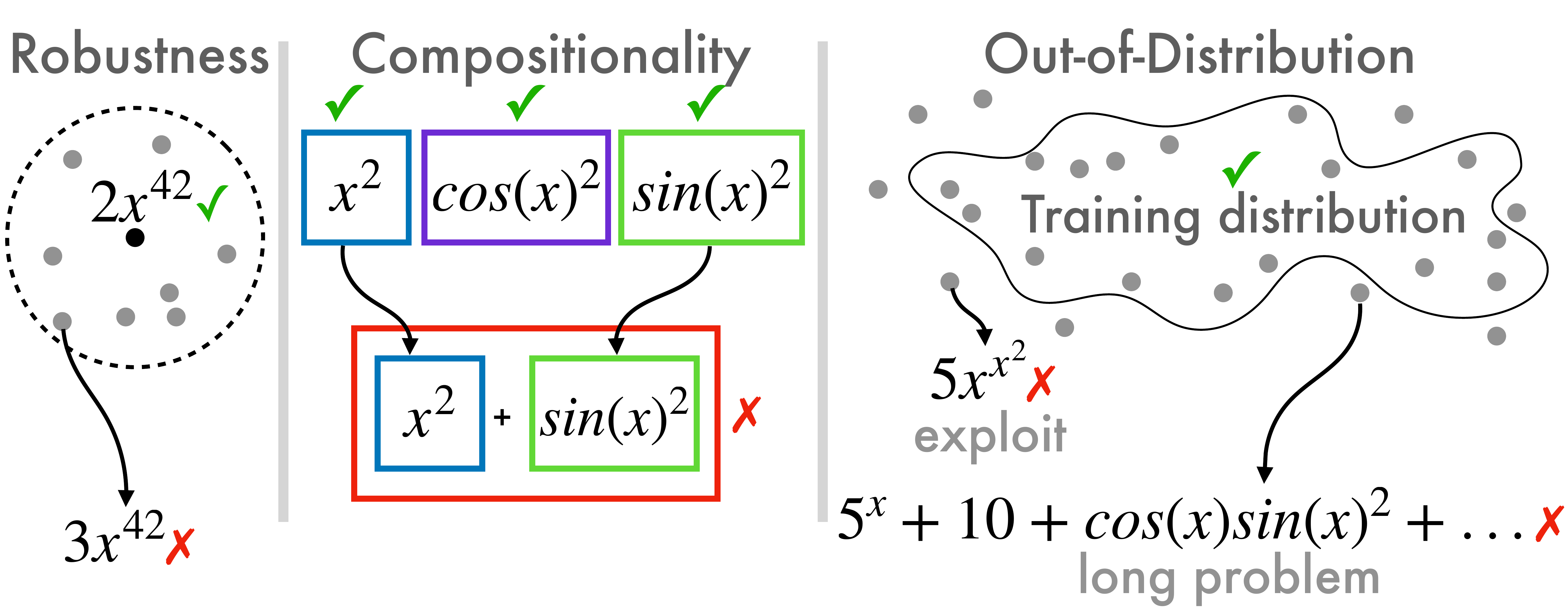}
\caption{\label{fig:fig1} Illustrating  \textit{robustness}, \textit{compositionality}, and \textit{out-of-distribution} deficiencies in the neural sequence integrator.
}
\end{center}
\end{figure}

In this paper, we find a discrepancy between the traditional notion of generalization captured by test set accuracy and the generalization needed in symbolic mathematics.
While the model's test accuracy is nearly perfect, we find this breaks down when testing its \textit{robustness}, \textit{compositionality}, and \textit{out-of-distribution generalization} (e.g. Table~\ref{tbl:front-examples}).
We describe a methodology for evaluating these aspects, by constructing problem sets and developing a genetic algorithm, SAGGA (Symbolic Archive Generator with Genetic Algorithms), that automatically discovers diverse and targeted failures.
We find that successfully integrating an in-distribution problem  does not imply success on nearby problems, despite being governed by the same underlying rule (\emph{robustness}).
Moreover, the model often succeeds on a collection of problems without being able to systematically compose those problems (\emph{compositionality}), and struggles to generalize to longer problems, larger values, and functions not covered in training (\emph{out-of-distribution}).

In addition to the model's approximate mode  being incorrect -- i.e. the most probable sequence returned by beam search -- the deficiencies are present deeper into its ranked list of candidate solutions, impacting the model's effectiveness in a \textit{search-and-verify} setting.
Overall, our investigation highlights the difficulty of achieving robustness, compositionality, and out-of-distribution generalization with the predominant modeling and learning approach, and the importance of evaluating beyond the test set, across aspects of generalization that are required by the task at hand.

\section{Problem Setup}
Symbolic integration is the problem of finding the integral $\yb$ of an input equation $\xb$.
For instance, $x^2/2$ is the integral of $x$, up to an additive constant.
\paragraph{Neural sequence integrator.}
\citet{lample2019deep} frame symbolic integration as a sequence-to-sequence problem.
In this view, input and output equations $\xb$ and $\yb$ are prefix-notation sequences.
The \textit{neural sequence integrator} uses a 6-layer transformer \citep{vaswani2017attention} to model the distribution $p_\theta(\yb|\xb)=\prod_{t=1}^{T_{\yb}}p_\theta(y_t|y_{<t},\xb)$ by training the model to maximize the log-likelihood of a set of training problems, $\arg\max_\theta \sum_{(\xb,\yb)\in \mathcal{D}} \log p_{\theta}(\yb|\xb)$.
Given a trained model and input $\xb$, a set of predicted solutions ranked by a model score is obtained by beam search, denoted $\{\hat\yb_1,\ldots,\hat\yb_k\}=f_\theta(\xb;k,b)$, where $b$ is beam size and $k$ is the number of candidates saved for evaluation.
For brevity we omit $b$ in the discussion unless necessary.

\paragraph{Evaluation.}
The standard practice is to evaluate a candidate $\hat\yb$ by checking whether the derivative of $\hat\yb$ is equivalent to $\xb$ using a symbolic solver (e.g. Sympy).
In the \textit{maximum-a-posteriori} (MAP) setting, the model's output is considered correct if its \textit{top-ranked} candidate $\hat\yb_1$ is correct.
This criterion is relaxed in the \textit{search-and-verify} setting, where the model's output is considered correct if \textit{any} of its $k$ candidates $\{\hat\yb_1,\ldots,\hat\yb_k\}$ is correct.
In this view, the neural network narrows the search space to a small set of candidates that are checked, trading off correctness for search and verification cost.
We denote checking $k$ candidate solutions as,
\begin{align}
\label{eqn:metric}
    m(\xb,f_\theta(\xb;k)) &= \begin{cases}
       0 & \xb \equiv \frac{d}{dx}\hat\yb_i \text{ for any } i\in 1 \text{ to } k,\\
       1 & \text{otherwise}.
    \end{cases}
\end{align}
In other words, $m(\cdot,\cdot)$ is 1 when the model \textit{fails} to predict the correct integral, and 0 when the model \textit{succeeds}.
We measure the proportion of failures on problems $X=\{\xb_1,\ldots,\xb_N\}$ using $k$ candidate solutions per problem as:
\begin{align}
    \texttt{Fail@k}(f_\theta,X)=\frac{1}{N}\sum_{\xb\in X} m(\xb,f_\theta(\xb;k)).
\end{align}
\texttt{Fail@k} is 0 when the model correctly integrates all of the problems in $X$, and increases towards 1 as it fails to integrate more problems.
Evaluating a model's performance in the MAP setting corresponds to evaluating $\texttt{Fail@1}$, while the search-and-verify setting with a budget of $k>1$ candidates uses $\texttt{Fail@k}$.
We omit $k$ in $f_\theta(\xb;k)$ unless necessary.

\subsection{Experiment Structure}
We structure our investigation into three parts (Figure~\ref{fig:fig1}).
We begin close to the model's training distribution, evaluating \textit{robustness} to small perturbations of in-distribution problems and simple functions.
We then ask whether learning to integrate a collection of functions implies that the model can integrate a \textit{composition} of those functions.
Finally we depart from the training distribution by studying \textit{extrapolation} to larger problems and values, then by finding adversarial \textit{exploits} that expose gaps in the training distribution.

\paragraph{Experimental setup.}
We use the implementation and pre-trained model from \citet{lample2019deep} for all of our experiments, specifically the \texttt{FWD+BWD+IBP} model which obtained top-10 accuracies of 95.6\%, 99.5\%, and 99.6\% on their publicly available test sets.\footnote{\url{https://github.com/facebookresearch/SymbolicMathematics/}, commit \texttt{4596d07}.}
Our evaluation is based on their code, we use their utilities for inputs and outputs, and by default use beam search with beam-size 10.
Following the authors, we use Sympy to check whether the derivative of a prediction is equal to the original problem.
We generously count the prediction as correct if a timeout occurs.
See the Apppendix for additional details.

\subsection{Automatic Problem Discovery with SAGGA}
Automatically finding problems that expose deficiencies requires a non-differentiable cost (Equation~\ref{eqn:metric}), satisfying constraints for valid equations, and finding diverse problem sets to characterize each aspect of generalization.
To address these challenges, we develop SAGGA (\textbf{S}ymbolic \textbf{A}rchive \textbf{G}eneration with \textbf{G}enetic \textbf{A}lgorithms), a gradient-free genetic algorithm which iteratively finds diverse failures.

At each iteration, SAGGA mutates a seed set of problems by modifying each problem's equation tree, ensuring that the resulting candidates are valid equations.
The candidates are scored by a fitness function -- i.e. according to whether the neural sequence integrator fails to integrate the problem and other desired constraints -- and the highest-fitness candidates are saved in a problem archive.
The next seed set is then formed to balance diversity and fitness, by clustering candidates and selecting the highest-fitness members of each cluster. 
SAGGA continues until the archive contains a target number of problems.
Algorithm~\ref{alg:sagga} summarizes SAGGA.

SAGGA offers control over the types of problems that it discovers through its seed problems, fitness function, and mutation strategy.
We detail our choices for each kind of generalization in their respective sections, and show default settings and further implementation details in the Appendix.

\begin{algorithm}[t]
\DontPrintSemicolon
\KwParams{Fitness $F(f_\theta, \xb)\rightarrow \mathbb{R}$,\\ \texttt{mutate} and \texttt{seed} strategies, archive size $N$.}
\KwOut{Problem archive $\mathcal{D}=\{\xb_1,\ldots,\xb_N\}$.}
$\mathcal{D}=\emptyset$ \tcp*[r]{initial archive}
$\hat{\xb}^{(0)}_{1:M}=\texttt{seed}(\mathcal{D},\emptyset)$ \tcp*[r]{initial seed}
\While{$|\mathcal{D}| < N$}{
\tcp*[l]{generate mutations} 
$\tilde\xb_{1:M'}^{(i)}=\texttt{mutate}(\hat{\xb}^{(i)}_{1:M})$\vspace{0.2em}\\
\tcp*[l]{select problems by fitness}
$\xb_{1:M''}^{(i)} = \texttt{select}(F, \tilde\xb_{1:M'}^{(i)}) $\vspace{0.2em}\\
\tcp*[l]{archive selected problems}
$\mathcal{D}=\mathcal{D} \cup \xb_{1:M''}^{(i)}$\vspace{0.2em}\\
\tcp*[l]{choose next seed} 
$\hat{\xb}^{(i+1)}_{1:M} = \texttt{seed}(\mathcal{D}, F, \tilde\xb_{1:M'}^{(i)})$
}
\caption{SAGGA. Each seed problem denoted as $\hat\xb$, mutated problem as $\tilde\xb$, archived problem as $\xb$.}
\label{alg:sagga}
\end{algorithm}

\section{Robust or Brittle?}
\label{sec:robustness}
First, we study whether the model's strong test-set performance adequately represents its \textit{robustness}.
Robustness tells us whether the integration model systematically solves all problems in a neighborhood governed by a generalizable pattern; for instance a model that solves $\int 26x^{42}$ should solve $\int 53 x^{42}$.
We study problems that are nearby to those from the original test distribution, as well as to simple primitive functions that offer fine-grained, interpretable control.

A \textbf{robust} model is stable to small perturbations in input, meaning that it gets nearby problems $\txb$ correct when it gets a problem $\xb$ correct.
Formally, let $X=\{\xb_1,\ldots,\xb_N\}$ contain problems that the model gets correct,
$\sum_{\xb\in X}m(\xb,f_\theta(\xb))=0,$
and let $\mathcal{N}_d(\xb)$ be a set of problems that are nearby to $\xb$ according to a distance $d(\xb,\txb)$.
We measure robustness by measuring failures on nearby problems,
\begin{align}
\label{eqn:robust-fail}
\texttt{Fail@k}(f_\theta, X_\mathcal{N}),
\end{align}
where $X_\mathcal{N}=\bigcup_{\xb\in X}\mathcal{N}_d(\xb)$.
We measure this quantity by varying (i) the neighborhood $\mathcal{N}_d(\xb)$ used to generate nearby problems, and (ii) the seed problems $X$ to consider.
Below, we will refer to a problem as $\xb$ or $f$ interchangeably.

\begin{table}[]
\setlength{\tabcolsep}{5pt}
\begin{center}
\begin{tabular}{llrrr}
\toprule
\textbf{Type} & \textbf{Test} & \textbf{Fail@50}& \textbf{Fail@10} & \textbf{Fail@1}  \\
\toprule
\texttt{coeff} & $k_1 \ln(k_2 x)$ & 0.0 & 0.0 & 0.0\\
               & $k_1 x$          & 0.0 & 0.0 & 0.0\\
               & $k_1 x^{42}$     & 0.0 & 6.1 & 45.5 \\
               & $k_1 \exp(k_2 x)$& 15.4 & 20.8 & 30.3 \\
               & $k_1 \sin(k_2 x)$& 6.6 & 19.6 & 29.7 \\
               & $k_1 \cos(k_2 x)$& 10.6 & 20.7 & 28.2 \\
               & $k_1 \tan(k_2 x)$& 13.9 & 17.4 & 24.2 \\
 \midrule
 $\texttt{coeff}$&$1/k\cdot f$ & 5.9 & 12.0 & 13.7 \\
 $\texttt{coeff}$ & $k\cdot f$ & 5.4 & 5.8  & 16.3 \\
 +$\texttt{exp}$ & $f + e^x$   & 0.9 & 1.6  & 3.3 \\
 +$\texttt{ln}$ & $f + \ln(x)$ & 1.9 & 3.2  & 5.3 \\
  \bottomrule
\end{tabular}
\caption{\label{tbl:robust-simple} \textit{Robustness} results with \textbf{simple primitives} (top) and \textbf{validation problems} $f$ which the model correctly integrates (bottom).
Coefficients are sampled from [1, 100].
}
\end{center}
\vspace{-1em}
\end{table}

\begin{table}[t!]
\setlength{\tabcolsep}{6pt}
\begin{center}
\begin{tabular}{lllr}
\toprule
\textbf{Input} & \textbf{Integral} & \textbf{Prediction} & \\
\midrule
$30\cos(39x)$ &  $\frac{10}{13}\sin(39x)$ & $\frac{10}{13}\sin(39x)$ & \cmark\\[1pt]
$17\cos(83x)$ & $\frac{17}{83}\sin(83x)$ & ${\color{red}{\frac{1}{17}}\sin(83x)}$ &\xmark\\[1pt]
$34\cos(77x)$ & $\frac{34}{77}\sin(77x)$ & ${\color{red}\sin(77x)}$ & \xmark\\
\midrule
$26x^{42}$ & $\frac{26}{43}x^{43}$& $\frac{26}{43}x^{43}$ & \cmark\\[1pt]
$88x^{42}$ & $\frac{88}{43}x^{43}$& ${\color{red}8x^{43}}$ & \xmark\\[1pt]
$53x^{42}$ & $\frac{53}{43}x^{43}$& ${\color{red}(x^{44} + x)/x}$ & \xmark\\
\bottomrule
\end{tabular}
\caption{\label{tbl:robustness-examples} 
\textit{Robustness} examples.
We show the model's top prediction (beam search, size 10).
Note that $(x^{44} + x)/x=x^{43}+1$; its derivative is $43x^{42}$ and is hence incorrect.
}
\end{center}
\vspace{-1em}
\end{table}

\subsection{Manually Testing Robustness}
To define nearby problems, we first
consider manual templates which minimally perturb a problem $f$, e.g.
\begin{align*}
k\cdot f, \quad f + \ln x, \quad \ldots
\end{align*}
These problems are nearby $f$ in the sense that a single operation is added to the problem's equation tree, or a small number of node values are changed in the tree.

\paragraph{Brittleness on simple primitive functions.}
We first investigate whether the neural sequence integrator is robust on \textit{simple primitive functions}, since they  make up more complicated functions and are frequently entered by real-world users.
We use a manual neighborhood which yields,
\begin{align*}
    X_{\mathcal{N}}=\{&k_1\ln (k_2 x),\quad k_1\exp(k_2 x), \quad k_1x, \quad k_1x^{42},\\ 
    & k_1\sin (k_2 x),\quad k_1\cos (k_2 x),\quad k_1\tan(k_2x)\},
\end{align*}
where $k_1 \sim \mathcal{U}(a, b)$ and $k_2\sim \mathcal{U}(a, b)$ are randomly sampled coefficients from a range $(a, b)$.
We use $[0,100]$ which is covered by the training distribution 
and  evaluate on 1,000 $k_1,k_2$ pairs sampled without replacement for each primitive.

Table~\ref{tbl:robust-simple} shows the results.
On a positive note, the neural sequence integrator is robust on the primitives $k_1 x$ and $k_1 \ln (k_2 x)$.
The integral of $k_1 x$ is $\frac{k_1}{2} x^2$, so the model learned to divide by 2 for these cases.
The integral of $\ln$ involves copying the coefficients into a correct template (that is, $\int k_1\ln(k_2 x)=k_1 x(\ln(k_2 x) - 1)$), and the neural sequence integrator  learned this behavior.

On the other hand, the model is surprisingly brittle on the other primitives.
These require dividing coefficients (e.g. $\int k_1\cos(k_2 x)=\frac{k_1}{k_2}\sin (k_2 x)$).
The failure rate shows that the model has not perfectly learned the required division behavior.
Moreover, despite learning a `division by 2' rule for integrating $k_1 x$, the neural sequence integrator's failures on $k_1 x^{42}$ indicate that it did not perfectly learn an analogous `division by 43' rule.
Table~\ref{tbl:robustness-examples} shows examples.

\paragraph{Test accuracy does not imply robustness.}
Next, we want to see whether the neural sequence integrator's strong test accuracy implies that it is robust on test problems.
We use the validation set, and perturb \textit{validation problems that the model correctly integrates} using the neighborhoods,
\begin{align*}
    X_{\mathcal{N}_1}=\{&\frac{1}{k} f,\quad k\cdot f\},\quad X_{\mathcal{N}_2}=\{f + e^x, \quad f+\ln(x)\},
\end{align*}
where $k \sim \mathcal{U}(1, 100)$.
The first set multiplies the function by a constant, while the second adds a single primitive.

Table~\ref{tbl:robust-simple} shows the results.
Despite achieving perfect accuracy on the original problems, the model frequently fails under the slight perturbations.
The local neighborhood around validation examples reveals deficiencies in robustness that are not evident from validation performance alone, aligning with findings in NLP tasks \citep{gardner2020evaluating}.

\subsection{Automatically Finding Robustness Failures}

Next, we use SAGGA to automatically discover robustness failures in the neighborhood of a seed set of problems.

\paragraph{Discovering brittleness near simple problems.}
First, we run SAGGA and only allow it to mutate leaves in a problem's equation tree into a random integer.
The problems are nearby in the sense that the tree's structure is not changing, only a small number of its leaf values.
We use SAGGA to mutate the leaves of seed sets of 9 polynomials $X_\texttt{poly}$ and 9 trigonometric functions $X_\texttt{trig}$, which are listed in the Appendix.
We run SAGGA until it discovers 1000 failing problems, then cluster these using k-means on SciBERT embeddings \citep{beltagy2019scibert} of each problem.

Table~\ref{tbl:sagga-robustness-examples} shows three members from three discovered problem clusters, for the polynomial and trigonometric seeds.
Intuitively, each cluster shows failures in a neighborhood around a prototypical problem -- for instance, on $2x^{42}+\textbf{k}$ the neural sequence integrator \textit{correctly} integrates $2x^{42}+\textbf{21}$, but not the problems in Cluster 2 (e.g. $2x^{42}+\textbf{22}$).
See Appendix Table~\ref{apx:tbl:sagga-robustness} for more prototypes and the model's predictions.

Curiously, each problem in a neighborhood is governed by a common template -- e.g. the problems $\{-104, -136, -33\}$ are governed by $\int k=kx+C$, yet the failures suggest that the neural sequence integrator has either not inferred the template, or does not apply it across the neighborhood.
To investigate this phenomenon, we show the \textit{raw} model prediction in Table~\ref{tbl:raw-outputs}, along with its simplified version and derivative.
Compared to the underlying template $\int k = kx+C$ the model's raw output is long and complex.
In contrast, the simplified version is short; we hypothesize this gap makes adhering to the template difficult.

\begin{table}[t!]
\setlength{\tabcolsep}{4pt}
\footnotesize
\begin{center}
\begin{tabular}{lccccc}
\toprule
\textbf{Seed} & \textbf{Cluster 1} & \textbf{Cluster 2} & \textbf{Cluster 3} \\
\midrule
$X_\texttt{poly}$& $-104$ & $2x^{42} + 22$ & $-47 + 2/x - 2/x^{71}$ \\
& $-136$ & $2x^{42} + 28$ & $-47 + 2/x - 31/x^{71}$ \\
& $-33$ & $2x^{42} + 68$ & $-71 + 36/x + 2/x^{71}$ \\
\midrule
$X_\texttt{trig}$ & $13\cos{19x}$ & $13\cos{83x} - 59$ & $10\sin{47x}\cos{2x}$ \\
& $13\cos{83x}$ & $17\cos{37x} - 49$ & $10\sin{90x}\cos{2x}$ \\
& $17\cos{47x}$ & $17\cos{41x} - 45$ & $19\sin{90x}\cos{2x}$ \\
\bottomrule
\end{tabular}
\caption{\label{tbl:sagga-robustness-examples} 
Example robustness problems discovered by SAGGA which the neural sequence integrator fails to integrate.
See Appendix Table~\ref{apx:tbl:sagga-robustness} for model predictions.
}
\end{center}
\end{table}

\begin{table}[]
    \centering
\begin{tabular}{llll}
\toprule
$\xb$ &  \textbf{Raw} & \textbf{Simplified} &   \textbf{Deriv.} \\
\midrule
$-104$ &$x^2 + 2x - (x + 25)^2$ &  $-48x - 625$ &   $-48$ \\
$-136$ &$x^2 - x(x + 130) + 2x$ &  $-128x$ &   $-128$ \\
$-33$  & $x^2 + x - (x + 16)^2$ &  $-31x - 256$ &   $-31$ \\
\bottomrule
\end{tabular}
    \caption{The raw model predictions for the problems $\xb$ and their simplified forms. Each prediction is incorrect since its derivative is not equal to $\xb$.
    The neural sequence integrator's raw output is long and varied compared to the underlying integration rule $\int k=k x + C$.
    }
    \label{tbl:raw-outputs}
\end{table}

\paragraph{Discovering brittleness near validation problems.}
Finally, we use SAGGA to discover difficult problems that are close to a target set $X$ -- in our case validation problems -- according to an explicit distance $d(\xb,\txb)$.
This allows for less hand-designing of the perturbations.

Specifically, we define a fitness which is high whenever a candidate is close to any problem in a target set $X$,
\begin{align}
\text{fitness}(\mathbf{\tilde x}) &=  \left[\min_{\mathbf{x}\in X}d(\mathbf{x},\mathbf{\tilde x})\right]^{-1}\cdot m(\tilde{\mathbf{x}}, f_\theta(\mathbf{\tilde x})).
\end{align}
We randomly sample 10 validation problems to form $X$, set SAGGA's initial seed to $X$,
and use cosine similarity of SciBERT vectors to define the distance $d(\xb,\txb)$.
Since the \textit{distance} now constrains the problems, we are free to use a wider set of mutations: changing a node's operation, adding an argument to a node, and replacing the node with a random constant, symbol, or simple operation.

Table~\ref{tbl:sagga-target-functions} shows example problems that SAGGA discovers around the successful validation problems, exposing a wider class of robustness failures than our preceding experiments.

\begin{table}[t!]
\setlength{\tabcolsep}{4pt}
\footnotesize
\begin{center}
\begin{tabular}{ll}
\toprule
\textbf{Validation Problem} & \textbf{Nearby Failures} \\
\midrule
$-x^2 + x + \log(4)\tan(x)$ & $-x^2 + x + \log(4)\tan(17x^2)$ \\
& $-x^2 + x + \log(4)\tan(2x^2)$ \\
& $-x^2 + x + \log(4)\tan(63/x^2)$ \\
\midrule
$\sqrt{3x + 3} - 2 $ & $\sqrt{-86^{x^2} + 62/x} - 40$ \\
& $\sqrt{14 + 62/x} + 4$ \\
& $\sqrt{14 + 62/x} - 2$ \\
\midrule
$\tan(\exp(2))/18x$ & $\tan(\exp(2 + 71/x))/18x$ \\
& $\tan(\exp(2 - 46^x))/18x$ \\
& $\tan(\exp(37x))/18x$ \\
\bottomrule
\end{tabular}
\caption{\label{tbl:sagga-target-functions} 
SAGGA discovers failures around successful validation problems, within a neighborhood defined by an explicit distance. 
Model predictions are in Appendix Table~\ref{apx:tbl:sagga-robustness}.
}
\end{center}
\end{table}

\section{Integrating Parts But Not The Whole}
\label{sec:compositionality}
While the preceding section identified weaknesses in robustness -- for instance, integrating $26x^{42}$ but not $88x^{42}$ -- a remaining question is whether successfully integrating a collection of primitives implies that a \textit{composition} of those primitives will be successfully integrated.

Compositionality refers to forming compound equations from known primitives and operations.
A compositional model should correctly integrate equations of the form,
\begin{align}
f=f_1 \circ f_2 \circ \cdots \circ f_k,
\end{align}
where $f_1,\ldots,f_k$ are equations that the model successfully integrates, and $\circ$ is a binary operation (e.g. addition).
For instance, a system that integrates $x^2$ and $\cos x$ and is capable of addition should successfully integrate $x^2 + \cos x$.

Formally, we say that a model is $k$-compositional with respect to equations $X$ and operation $\circ$ when it successfully integrates any combination of $k$ equations from $X$,
$\sum_{\xb\in \tilde{X}} m(\xb,f_\theta(\xb))=0$,
where $\tilde{X}=\{f_1\circ\cdots \circ f_k |f_i\in X\}$.

We evaluate $k$-compositionality with respect to addition, using simple primitive functions and validation problems.
As integration is linear, $\int (f+g)=\int f + \int g$, compositionality with respect to addition is a reasonable requirement.

\begin{table}[t!]
\setlength{\tabcolsep}{5pt}
\begin{center}
\begin{tabular}{lllr}
\toprule
\textbf{Input} & \textbf{Prediction} & \\
\midrule
$x^{1/3}$ & $\frac{3}{4}x^{4/3}$ & \cmark\\[1pt]
$x^{1/606}$ & $\frac{606}{607}x^{\frac{607}{606}}$ & \cmark\\[1pt]
$x^{1/3}+x^{1/606}$ &${\color{red}{\frac{3}{5}x^{\frac{5}{3}}+\frac{6}{613}x^{\frac{613}{606}}}}$ &\xmark\\[2.5pt]
\hline 
\rule{0pt}{1.05\normalbaselineskip}$x^{209}$ & $\frac{1}{210}x^{210}$ & \cmark\\[1pt]
$x^{764}$ & $\frac{1}{765}x^{765}$ & \cmark\\[1pt]
$x^{209}+x^{764}$& ${\color{red}{\frac{1}{205}x^{205}}}$ &\xmark\\[2.5pt]
\hline 
\rule{0pt}{1.05\normalbaselineskip}$14\cos(58x)$ & $\frac{7}{29}\sin(58x)$ & \cmark\\[1pt]
$46\cos(84x)$& $\frac{23}{42}\sin(84x)$ & \cmark\\[1pt]
$14\cos(58x)+46\cos(84x)$ & ${\color{red}{\sin(59x)\cos(x)}}$ &\xmark\\
\bottomrule
\end{tabular}
\caption{\label{tbl:compositionality-examples} 
\textit{Compositionality} examples.
We show the model's top prediction (beam search, width 10).
The model successfully integrates the individual primitives, but not their sum.
}
\end{center}
\end{table}
\begin{table}[t!]
\setlength{\tabcolsep}{4pt}
\begin{center}
\footnotesize
\begin{tabular}{llrrr}
\toprule
\textbf{Type} & \textbf{Test}& \textbf{Fail@50} & \textbf{Fail@10} & \textbf{Fail@1}  \\
\midrule
$\texttt{exp(1)}$ & $f_1$            & 00.0 & 00.0 & 00.0 \\
$\texttt{exp(2)}$ & $f_1+f_2$        & 70.8 & 72.4 & 84.9 \\
$\texttt{exp(3)}$ & $f_1+f_2+f_3$    & 91.3 & 97.5 & 99.5 \\
$\texttt{exp(4)}$ & $f_1+\ldots+f_4$ & 86.2 & 97.4 & 99.8 \\[3pt]
\hline
\rule{0pt}{1.1\normalbaselineskip}$\texttt{coeff(1)}$ & $f_1 $ & 00.0 & 00.0 & 00.0\\
$\texttt{coeff(2)}$ & $f_1+f_2 $         & 8.60 & 16.2 & 29.2 \\
$\texttt{coeff(3)}$ & $f_1+f_2+f_3$      & 23.8 & 37.5 & 61.0 \\
$\texttt{coeff(4)}$ & $f_1+\ldots+f_4$   & 23.1 & 38.7 & 60.0 \\[3pt]
\hline
\rule{0pt}{1.1\normalbaselineskip}$\texttt{valid(1)}$ & $f_1$ & 00.0 & 00.0 & 00.0\\
$\texttt{valid(2)}$ & $f_1+f_2$                       & 6.80 & 14.5 & 15.0 \\
$\texttt{valid(3)}$ & $f_1+f_2+f_3$                   & 21.5 & 36.5 & 43.6\\
$\texttt{valid(4)}$ & $f_1+\ldots+f_4$                & 52.5 & 69.0 & 79.2\\
\bottomrule
\end{tabular}
\caption{\label{tbl:compositional} \textit{Compositionality}. Top:  \textbf{successful simple} primitives from the robustness experiments (Table~\ref{tbl:robust-simple}). Bottom: \textbf{successful validation-set} primitives. 
Despite integrating each primitive, the model struggles to integrate their sums.
}
\end{center}
\end{table}

\paragraph{Succeeding on simple primitives, failing on their sum.} 
We collect simple primitives from the coefficient robustness experiments that the model successfully integrates (\texttt{coeff}), and successful exponents $x^c$ or $x^{1/c}$, $c\in [0,1000]$ (\texttt{exp}).
We randomly sample 1000 compound equations $f_1+\ldots+f_k$ for $k\in \{2,3,4\}$ and evaluate the failure rate.
Table~\ref{tbl:compositional} shows the results.
Adding two primitives gives failure rates of 29\% and 85\% for coefficient and exponent primitives, respectively, despite failing 0\% of the time on the individual primitives.
As the number of compositions increases, the failure rate increases towards 100\%.
Table~\ref{tbl:compositionality-examples} shows examples.

\paragraph{Succeeding on test problems, failing on their sum.}
We perform a similar experiment using successful validation-set functions.
We filter examples longer than 20 tokens so that composed equations are within the training domain in terms of length, and sample 1000 compound equations $f_1+\ldots+f_k$ for $k\in \{2,3,4\}$.
As seen in Table~\ref{tbl:compositional}, the failure rate grows as the number of compositions increases, similar to the simple primitives case.
Maximizing the likelihood of a large training set did not yield a compositional model.

\section{Departing Further From Training}
\label{sec:ood}

The preceding experiments found problems that were nearby to, or composed directly from, in-distribution examples.
In this section, we deliberately move from the model's training distribution to evaluate its \textit{out-of-distribution} generalization.
First, we study \textit{extrapolation} to longer equation sizes than those in its training distribution, and to integer ranges that are only sparsely covered in the training set.
Then we use SAGGA to expose exotic failures and reveal problem classes that were not covered during training.

\begin{table}[t!]
\footnotesize
\setlength{\tabcolsep}{6pt}
\begin{center}
\begin{tabular}{crr}
\toprule
\textbf{Nodes} & \textbf{Fail@10}  &\textbf{Fail@1} \\
\midrule
 1-15 & 0.4 & 1.6  \\
 20   & 1.9 & 10.7  \\
 25   & 7.2 & 17.2  \\
 30   & 24.4 & 37.1  \\
 35   & 49.0 & 59.2 \\
\bottomrule
\end{tabular}
\caption{\label{tbl:extrapolation} \textit{Extrapolation} to more operator nodes under the training data generation process.
Training used 1-15 nodes.
}
\end{center}
\vspace{-1.5em}
\end{table}

\begin{figure}
\begin{center}
\includegraphics[width=0.95\columnwidth]{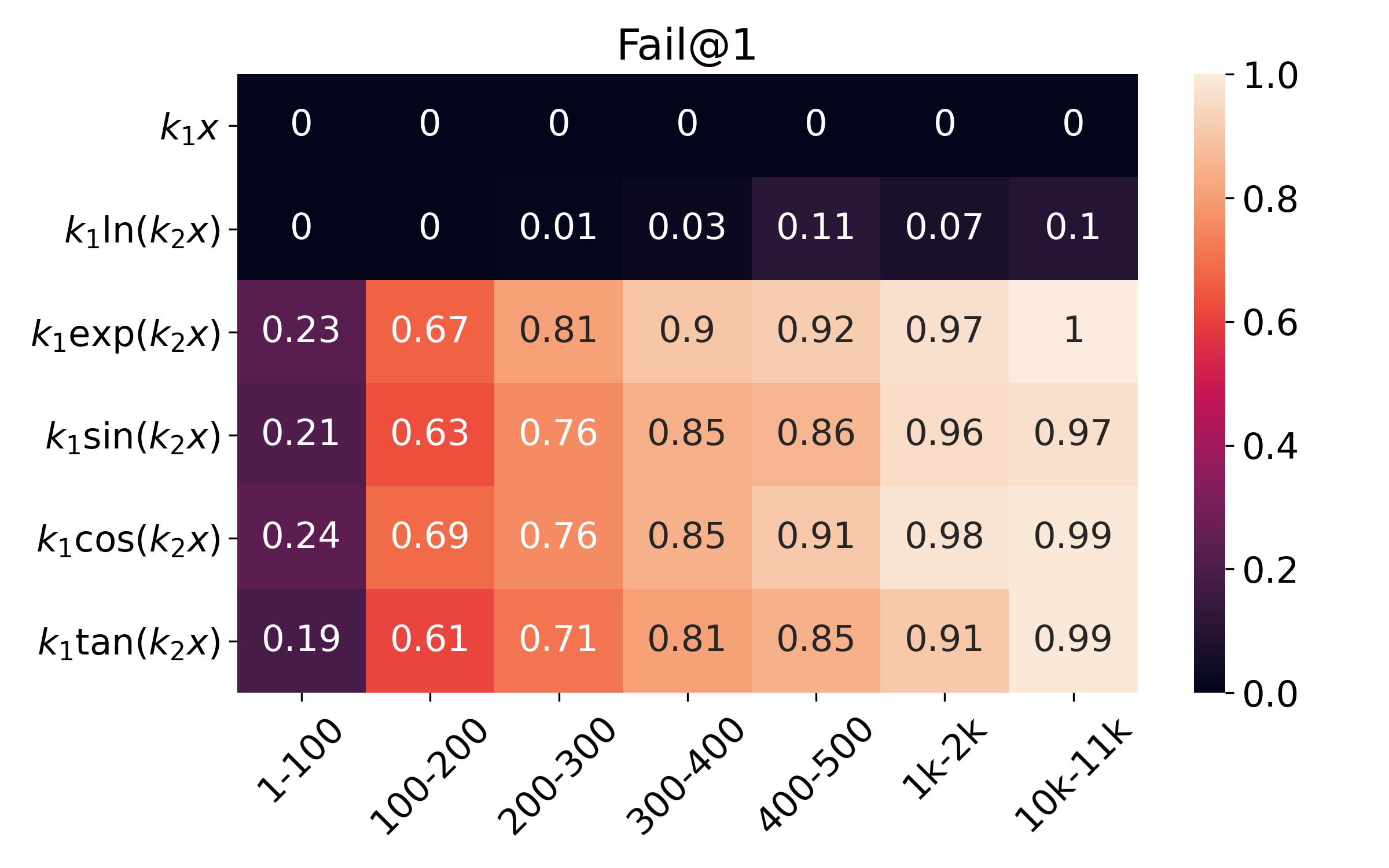}
\caption{\label{fig:integer-extrapolation} \textit{Integer extrapolation.} Failure rates for integrating simple primitives with coefficients from the specified range.}
\end{center}
\vspace{-1.5em}
\end{figure}

\paragraph{Longer problems are more difficult.}
First, we use the same data-generating process as for training, but \textit{vary its parameters} to depart from the training distribution.
Specifically, we test extrapolation on number of operator nodes in each equation tree, using \citet{lample2019deep}'s data generation process
and varying the \texttt{max\_ops} parameter.
Table~\ref{tbl:extrapolation} shows performance when \texttt{max\_ops} is increased past the model's training domain (1-15).
The neural sequence integrator does show some extrapolation to equation trees with more operator nodes than it was trained on, but its failure rate increases substantially as the number of nodes increases.

\paragraph{Larger failures on larger digits.} Next, we study performance as integer values increase, quickly going out-of-domain.
Considering a sample of 200,000 sequences from the training distribution, 99.4\% of the positive integers were between 1 and 100. Other ranges were non-empty but sparsely represented; for instance, 0.2\% of the integers were between 100 and 200, and 0.096\% between 1,000 and 2,000.
Figure~\ref{fig:integer-extrapolation} shows performance on primitive functions with coefficients from the specified range.
As in the robustness experiments, the $x$ and $\ln$ primitives perform well, showing that there is \textit{some} ability to use large numbers.
However, performance severely degrades for the $\texttt{exp}, \texttt{sin}, \texttt{cos}, \texttt{tan}$ primitives as the coefficient magnitudes increase, reaching near 100\% failure rates on large coefficients.

\begin{table}[t!]
\setlength{\tabcolsep}{4pt}
\footnotesize
\begin{center}
\begin{tabular}{cc}
\toprule
\textbf{Problem} & \textbf{Exploit}  \\
\midrule
 $169\sin(4x)/x$ & Uses $\mathrm{Si}(\cdot)$.  \\
 $-2\sin(42/x)$ & Uses $\mathrm{Ci}(\cdot)$.\\
 $-2\sin(185x^2)\cos(2x)$ & Uses Fresnel $\mathrm{S}, \mathrm{C}$ integrals.\\
 ${357^{x^2}}^x + 2\sin(2x)$ & Uses incomplete gamma $\Gamma(a,x)$.  \\
 $1/(x^{48}(3x+2)^{49})$ & Decoding does not terminate.\\
\bottomrule
\end{tabular}
\caption{\label{tbl:genetic-general-exploits} \textit{Exploits} discovered by SAGGA whose integrals use out-of-domain functions. See Appendix Table~\ref{apx:tbl:sagga-exploits} for model predictions.
}
\end{center}
\end{table}
\begin{table}[t!]
\setlength{\tabcolsep}{4pt}
\footnotesize
\begin{center}
\begin{tabular}{ccccc}
\toprule
 \textbf{Cluster 1} & \textbf{Cluster 2} & \textbf{Cluster 3} & \textbf{Cluster 4}\\
\midrule
$119^x$ & ${-240}^x + 2\cos{2x}$ & ${-100x}^x$ & ${158x}^{x^2} + 611$  \\
$132^x$ & ${-398}^x + 2\cos{2x}$ & ${-149x}^x$ & ${256x}^{x^2} + 191$  \\
$136^x$ & ${-692}^x + 2\sin{2x}$ & ${-151x}^x$ & ${332x}^{x^2} + 559$  \\
\bottomrule
\end{tabular}
\caption{\label{tbl:sagga-general} 
SAGGA discovers many failures that involve $x$ in an exponent.
See Appendix Table~\ref{apx:tbl:sagga-exponent} for model predictions.
}
\end{center}
\end{table}

\paragraph{Discovering unsupported functionality.}
Next, we run SAGGA in an unconstrained setting with all mutation types, favoring short problems using the fitness,
$F(f_\theta, \xb) = m(\xb,f_\theta(\xb))\cdot \frac{1}{|\xb|},$
which is positive when the model returns an incorrect integral for $\xb$, and higher for shorter problems.

SAGGA discovers \emph{exploits} based on the neural sequence integrator's limited training distribution, such as problems whose integral is expressed using the Gamma function $\Gamma(\cdot)$, or the cosine integral $\mathrm{Ci}$, which are not included in its training data (Table~\ref{tbl:genetic-general-exploits}).\footnote{\url{https://en.wikipedia.org/wiki/Trigonometric_integral}.}
These examples are a reminder that the sequence-to-sequence paradigm determines which functions are `built in' by inclusion in training data; omitted behavior is left unspecified, leaving it susceptible to exploits.

Finally, the last problem in Table~\ref{tbl:genetic-general-exploits} caused the neural sequence integrator to enter a non-terminating loop  during decoding (Appendix Table~\ref{apx:tbl:sagga-exploits}), a known idiosyncrasy of autoregressive models with beam search \citep{welleck2020consistency}.

SAGGA also finds many clusters that indicate the neural sequence integrator struggles when $x$ appears in an exponent.
The discovered problems in Table~\ref{tbl:sagga-general} are a microcosm of our previous findings:
For the first cluster, we manually found a nearby problem, $30^x$, that the model gets correct; this cluster is a \textit{robustness} failure.
The second cluster shows how such failures cascade further as the function is \textit{composed}.
The final two clusters involve $x^x$ or $x^{x^2}$, which do not have analytical integrals;\footnote{\url{https://www.wolframalpha.com/input/?i=integral+of+x**x}} these clusters are \textit{exploits}.

\begin{figure}
\begin{center}
\includegraphics[width=\columnwidth]{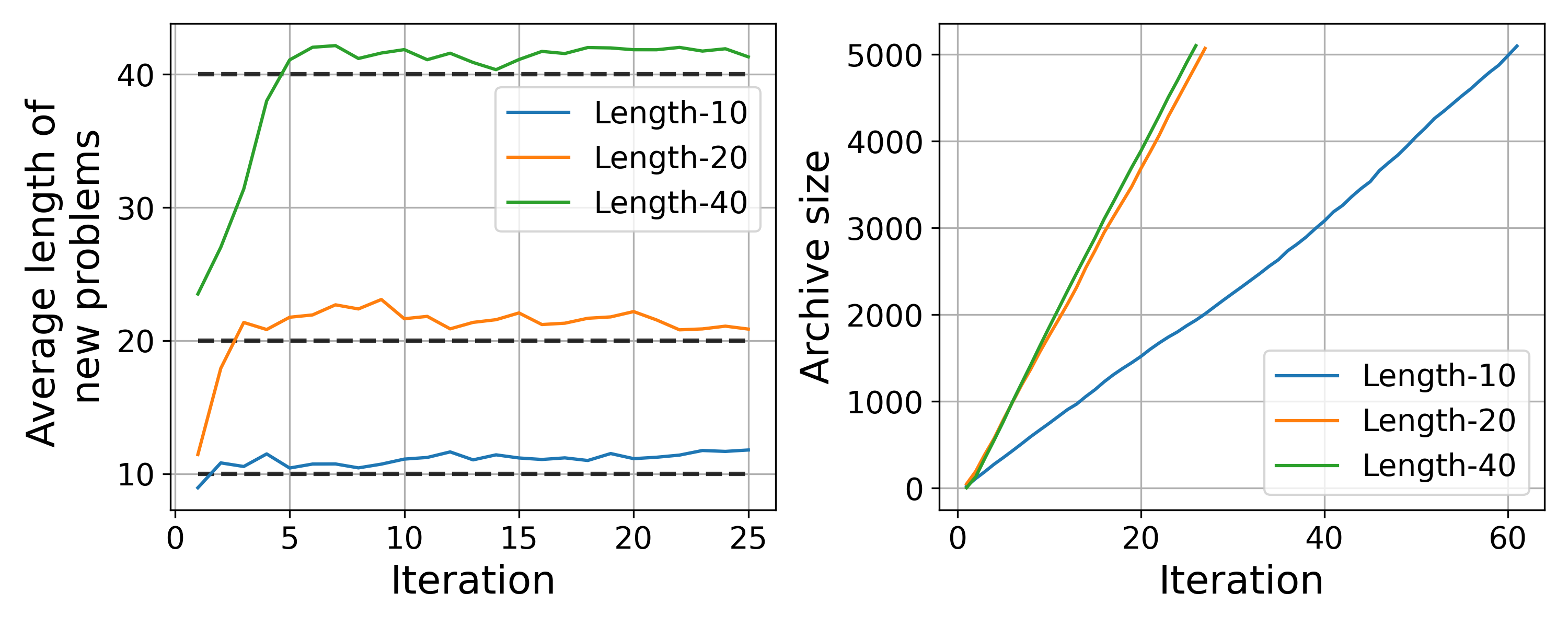}
\caption{\label{fig:genetic-target-length} Left: Average length of discovered problems in each iteration, using fitness functions that promote the given length.
SAGGA discovers problems of each target length.
Right: SAGGA discovers longer problems at a higher rate.
}
\end{center}
\end{figure}

\paragraph{Finding problems with target properties.} Finally, we generate failures of a target length, by running SAGGA to target length 10, 20, and 40 problems.
As seen in Figure~\ref{fig:genetic-target-length}, SAGGA converges to find problems of the target length.
Based on our extrapolation experiments, we expect SAGGA to fail more often on longer equations. 
The right-hand plot in Figure~\ref{fig:genetic-target-length} shows that it is also easier to \textit{find} failures for longer equations, in that the archive grows more quickly for longer target lengths.
While we visually inspect short equations for interpretability, the growth rate is a reminder that the space of failures is vast for longer equations.

\section{Is it a search problem? Distinguishing between model and search errors}

\begin{table}[t]
\begin{tabular}{llll}
\toprule
                & \textbf{$p$@1} & \textbf{$p$@500} & \textbf{$p$@\{1-500\}}  \\
\midrule
 Failures@500 &  0.93912 & 3.9$\times 10^{-6}$ & 0.99435 \\
 Success@500  &  0.91647 & 3.5$\times 10^{-6}$ & 0.99316\\
\bottomrule
\end{tabular}
\caption{Probability assigned to the top candidate ($p$@1), the 500'th candidate ($p$@500), and the mass covered by the top 500 candidates ($p$@\{1-500\}), in failures and successful cases (\texttt{exp}-robustness problems, width-500 beam search).
Sequences outside of the top 500 are very improbable compared to the most probable candidate.
In both cases, the top 500 candidates cover most of the probability mass.
}
\label{tbl:mass}
\end{table}

\begin{table}[t!]
\setlength{\tabcolsep}{6pt}
\begin{center}
\begin{tabular}{cccc}
\toprule
\textbf{k} &  \textbf{Unresolved@k} \\
\midrule
1  & 91.6\%  \\
10 & 65.2\%  \\
\bottomrule
\end{tabular}
\caption{\label{tbl:search} Percentage of failures on the \texttt{FWD} validation set in which the ground truth $\yb_*$ is scored lower than the top beam candidate (Unresolved@1) or the bottom beam candidate (Unresolved@10), meaning that perfect search would leave the failures at level $k$ unresolved.
}
\end{center}
\vspace{-1.5em}
\end{table}

Both the experiments of \citet{lample2019deep} and our own generate candidate solutions from a sequence-to-sequence model using beam search. 
This raises the possibility that failures are due to search rather than the model: what if the highest scoring sequences are correct, but not found?

Specifically, we want to distinguish between \textbf{search errors}, which occur when $p_\theta(\yb_*|\xb)\gg p(\yb|\xb)$ but the search algorithm (e.g. beam search) does not return $\yb_*$, and \textbf{model errors}, meaning $p_\theta(\yb|\xb)\gg p(\yb_*|\xb)$.
Here $\yb_*$ is any correct solution to problem $\xb$.

\subsection{The model is deficient: model errors.}
We study simple-robustness and in-distribution problems, and find evidence of model deficiencies that would remain unresolved with perfect search.

\paragraph{Robustness.} First, we study the simple-primitive robustness problems (e.g. $k_1\exp(k_2 x)$, see \autoref{tbl:robust-simple}), as these short problems resulted in a small number of  timeouts, allowing us to scale up search and verification.
We increase the beam size to 500 candidates, and study the model's probabilities on the 500 returned candidates and correct solutions.
We refer to the candidates ranked by decreasing probability as $\yb_{\text{beam}}^{(1)},\ldots,\yb_{\text{beam}}^{(500)}$ (i.e. $\yb_{\text{beam}}^{(1)}$ has the highest probability).

\begin{center}
\begin{figure}
    \includegraphics[width=0.9\columnwidth]{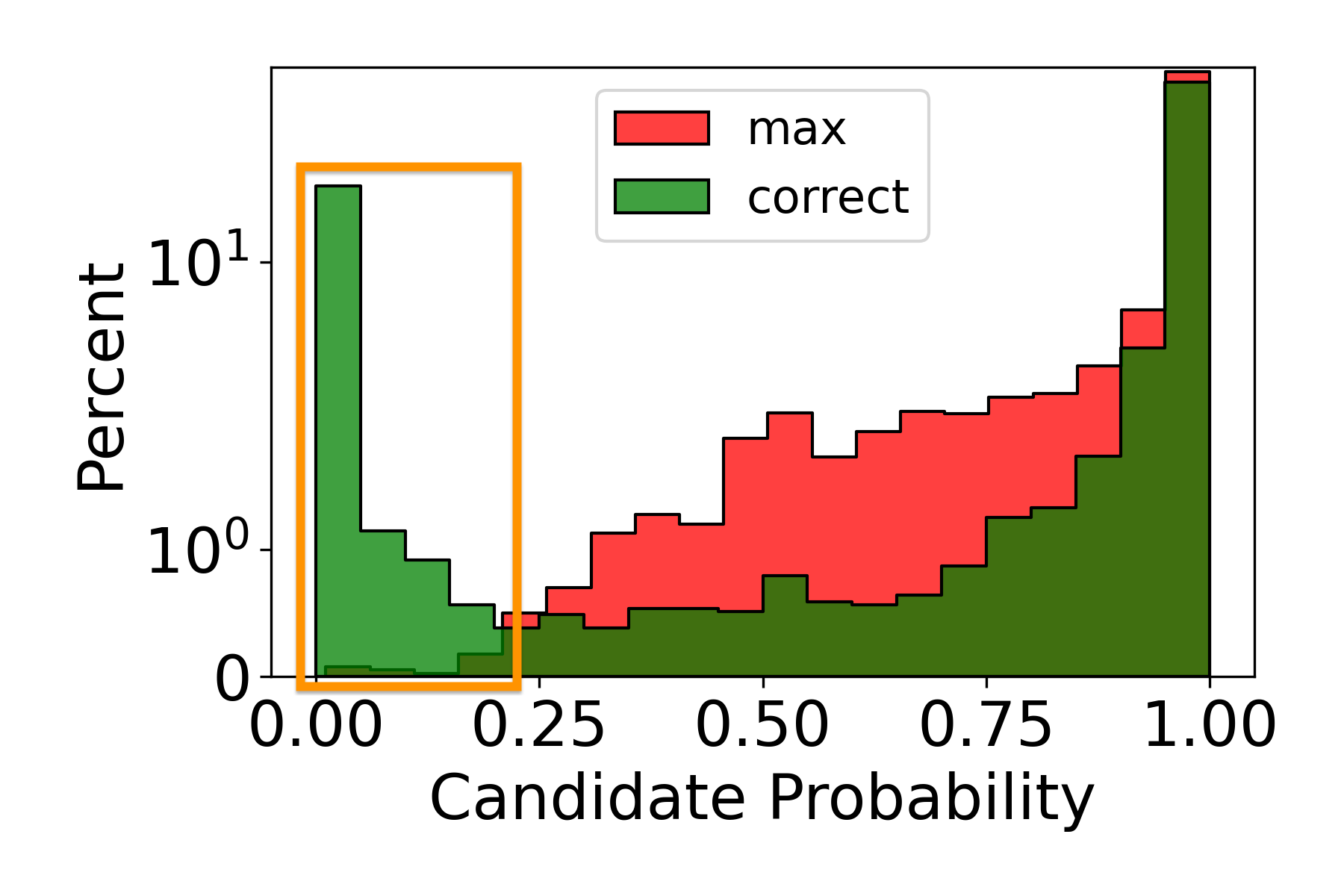}
    \vspace{-1em}
    \caption{Probabilities assigned to the top-ranked beam candidate (\texttt{max}) versus correct candidates (\texttt{correct}) with a large search \& verification budget (500 candidates). 
    The model often assigns very low probability to correct candidates that are found by increasing the search budget; search covers up underlying model deficiencies. (\texttt{exp}-robustness)
    }
    \label{fig:beam500-mismatch}
\end{figure}
\end{center}

When a correct solution is within the 500 returned candidates, the correct solution often has \textit{much lower probability} than the top candidate, $p_\theta(\yb_{\text{beam}}^{(1)}|\xb)\gg p(\yb_*|\xb)$.
Specifically, correct solutions often appear at the bottom of the candidates (\autoref{fig:beam500-mismatch}, orange), yet on average the bottom candidate $\yb_{\text{beam}}^{(500)}$ has probability $\approx 0.0000035$, while  the top candidate $\yb_{\text{beam}}^{(1)}$ has probability $\approx 0.92$ (\autoref{tbl:mass}).
These are model deficiencies: the model is \textit{confidently incorrect}, assigning very high probability to an incorrect solution at the top, and very low probability to correct solutions.

When a correct solution is \textit{not} within the top 500 candidates, the model is again confidently incorrect, with the top candidate $\yb_{\text{beam}}^{(1)}$ receiving $\approx 0.94$ probability.
Improving the search algorithm -- e.g. by further increasing the search budget or using an alternative to beam search -- would inevitably return a low probability solution, as the 500 candidates already cover more than $99.4\%$ of the probability mass (\autoref{tbl:mass}).
The findings again point to model errors.

\paragraph{In-distribution.} Next, we study \textit{in-distribution} problems from the \texttt{FWD} validation set of \citet{lample2019deep}.
On failure cases, we test if the ground-truth $\yb_*$ is scored above the top $k$ beam candidates, meaning that failure@$k$ might be resolved with perfect search--$\yb_*$ was scored more highly but was simply not found.
As seen in \autoref{tbl:search}, the majority of failures -- 91.6\% for failure@1 and 65.2\% for failure@10 -- would remain unresolved with perfect search, again pointing to model deficiencies.

\subsection{Covering up deficiencies with search.}

Our preceding experiments showed that the top of the model's ranked solution list is often made up of incorrect solutions, while correct solutions deeper into the solution list are assigned very low probabilities.
A natural question is whether we can simply `cover up' the deficient model by enumerating and verifying even more candidates, while ignoring the model's probabilities.

On simple robustness problems (e.g. $k_1\exp(k_2 x)$), we find that large search budgets can alleviate failures, i.e. Fail@$k$ decreases as $k$ increases, 
for instance moving from roughly 30\% failure@1 to 3\% failure@500 on $\texttt{exp}$ robustness (\autoref{apx:fig:beam500}).
On more complex problems, verifying more candidates reduces the failure rate, yet our experiments do not indicate that failures approach zero for practical search budgets.
For instance, in our compositionality $\texttt{exp}$ experiments (Table~\ref{tbl:compositional}), 
verifying 50 instead of 1 candidate reduces failures, but not below 90\%.
Larger search budgets quickly become impractical to verify: 
for compositionality $\texttt{exp}$,
the verification time increases substantially, from around 5 minutes for 1 candidate to over 2 hours for 50, with worst-case verification time of 41.6 hours (Table~\ref{tbl:runtime}).
Looking ahead, developing methods that decrease search and verification cost can help to further cover up a subset of model errors, yet improving the underlying model remains a core issue.

\begin{table}[t]
\begin{center}
\begin{tabular}{ccc}
\toprule
\textbf{Candidates} & \textbf{Verify (hrs)} & \textbf{Worst-case (hrs)} \\
\midrule
                  1 &                         0.088 &                       0.833 \\
                 10 &                         0.463 &                       8.333 \\
                 50 &                         2.250 &                      41.667 \\
\bottomrule
\end{tabular}
\caption{Runtime needed to verify the 3,000-problem exponent compositionality cases ($\texttt{exp}$, Table~\ref{tbl:compositional}) with \texttt{Sympy}, using $k$ candidates  for $k\in \{1, 10, 50\}$.}
\label{tbl:runtime}
\end{center}
\end{table}

\section{Related Work}

In this work, we study systematic generalization in sequence models applied to symbolic integration, in terms of robustness, compositionality, and extrapolation, and develop a genetic algorithm for building adversarial problem sets.

\paragraph{Symbolic mathematics and sequence models.}
Several works study extrapolation to longer sequences and larger digits in synthetic arithmetic and basic mathematics tasks \citep{zaremba2014,trask2018neural,saxton2019analysing,nogueira2021InvestigatingTL}.
Sequence models have also been applied to polynomial rewriting \citep{piotrowski2019can,agarwal2021analyzing}, and differential system stability \citep{charton2021learning}.
For symbolic integration, \citet{davis2019use} argue that the neural sequence integrator's test performance should be qualified, though without an empirical demonstration.
These critiques motivate our focus on the neural sequence integrator \citep{lample2019deep}, whose performance we characterize and empirically study in terms of systematic generalization.

\paragraph{Systematic generalization.}
Several works identify difficulties with modern methods on synthetic tasks (e.g. \citet{lake2018still,bahdanau2019,hupkes2020,kim2020cogs}) and machine translation \citep{raunak2019On}, with a focus on compositionality and extrapolation.
Some methods address systematicity with inductive biases in model structure \citep{andreas2016neural,bahdanau2019}, and others through the data \citep{hill2020environmental,andreas2020good} or learning procedure \citep{lake2019compositional,vani2021iterated}.
We focus on systematic generalization deficiencies in a state-of-the-art model in a new setting -- symbolic integration -- with additional aspects of generalization.

\paragraph{Robustness and adversaries in sequence models.}
Several works study robustness in NLP, including classification \citep{tu-etal-2020-empirical}, word substitutions \citep{jia-etal-2019-certified}, domain shift in QA \citep{kamath-etal-2020-selective} or topic distributions \citep{oren-etal-2019-distributionally}.
Several methods find adversarial examples in NLP  \citep{morris2020textattack}.
\citet{alzantot2018generating} use genetic algorithms in a classification setting, while we consider generation.
\citet{michel2019evaluation} constrain input sequences to be similar
and use a gradient-based attack to swap tokens. We face a non-differentiable cost and generate large collections of failures with a wide class of mutations.

\section{Conclusion}
We study generalization in symbolic mathematics using the predominant modeling paradigm: a large-scale transformer trained with maximum likelihood.
We find deficiencies that are not captured by test accuracy, including brittleness to small perturbations, difficulty composing known solutions, and gaps in the training distribution.
We offer speculations based on our results.
Due to the large space of equations, practical empirical distributions do not provide a dense sampling of individual problem types (e.g. $k_1\cos(k_2x)$), and each empirical sample contains shared biases from the underlying data generator (e.g. integer values, lengths).
Thus, sparse test sets do not adequately measure systematic generalization.
From a learning perspective, generic networks trained with SGD do not necessarily favor the simplest hypothesis to explain the data; thus a sparse training set yields an underconstrained hypothesis space, with hypotheses that do not strongly generalize (e.g. Table~\ref{tbl:raw-outputs}), causing behavior that breaks simple rules (e.g. adhering to a template or following the sum rule).
We suspect that inductive biases-- e.g. encoded through the training distribution, architectural components, or learning algorithm-- are needed to narrow the hypotheses to those that strongly generalize.

\bibliography{custom}

\clearpage
\appendix

\section{Additional Results}

\subsection{Large Search Budgets}

\begin{figure}
    \centering
    \includegraphics[width=0.75\columnwidth]{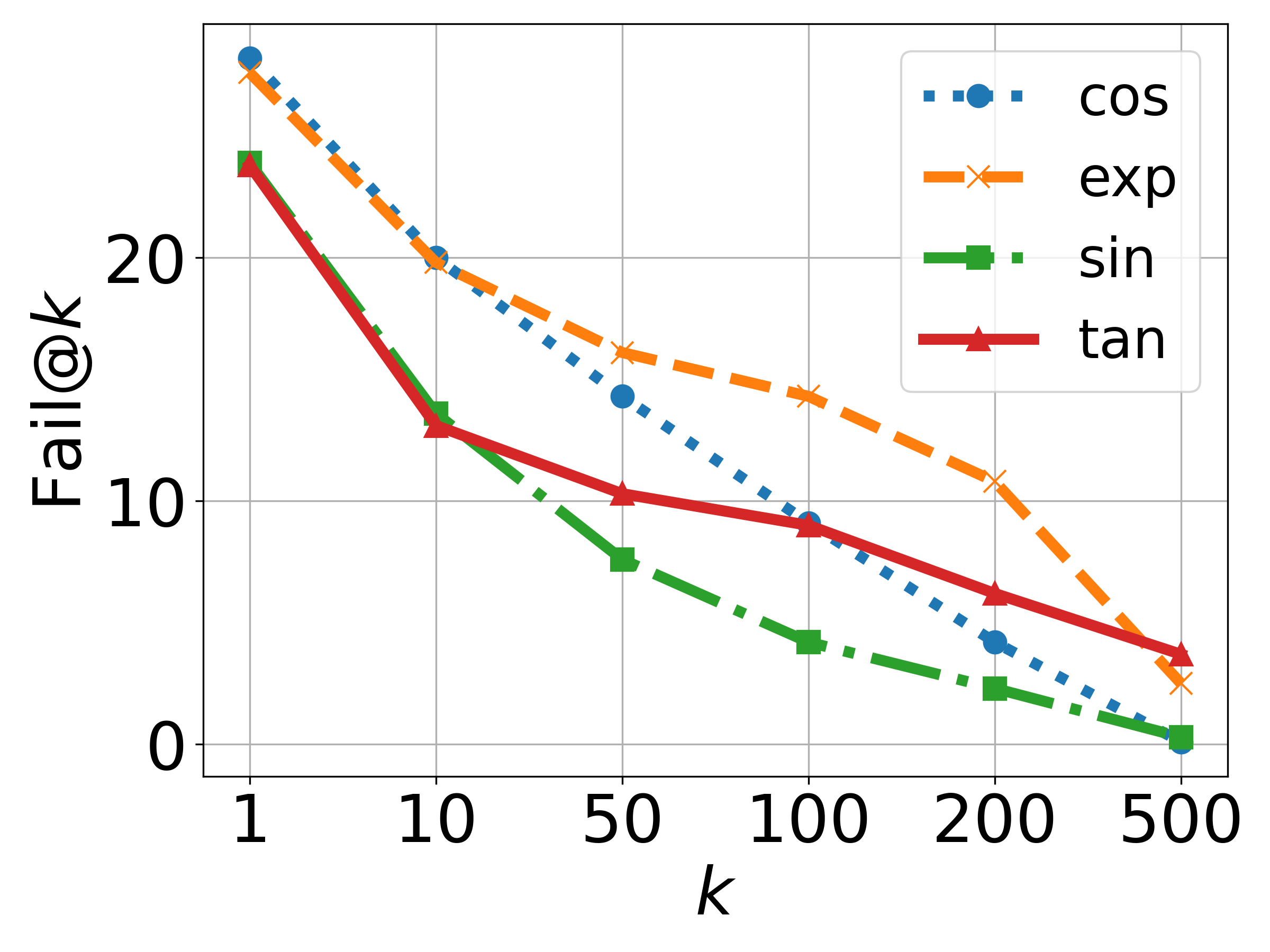}
    \caption{\texttt{Fail@k} with a \textbf{large search-and-verify budget} (500 candidates, beam search) on \textbf{simple-primitive robustness} problems.
    The model is `brittle' in that the top of its ranked solution list is often made up of incorrect solutions; e.g. the top-ranked solution is incorrect around 25\% of the time. 
    However, the correct solution often exists within the model's top 500 predictions on these problems.
    }
    \label{apx:fig:beam500}

    \includegraphics[width=0.6\columnwidth]{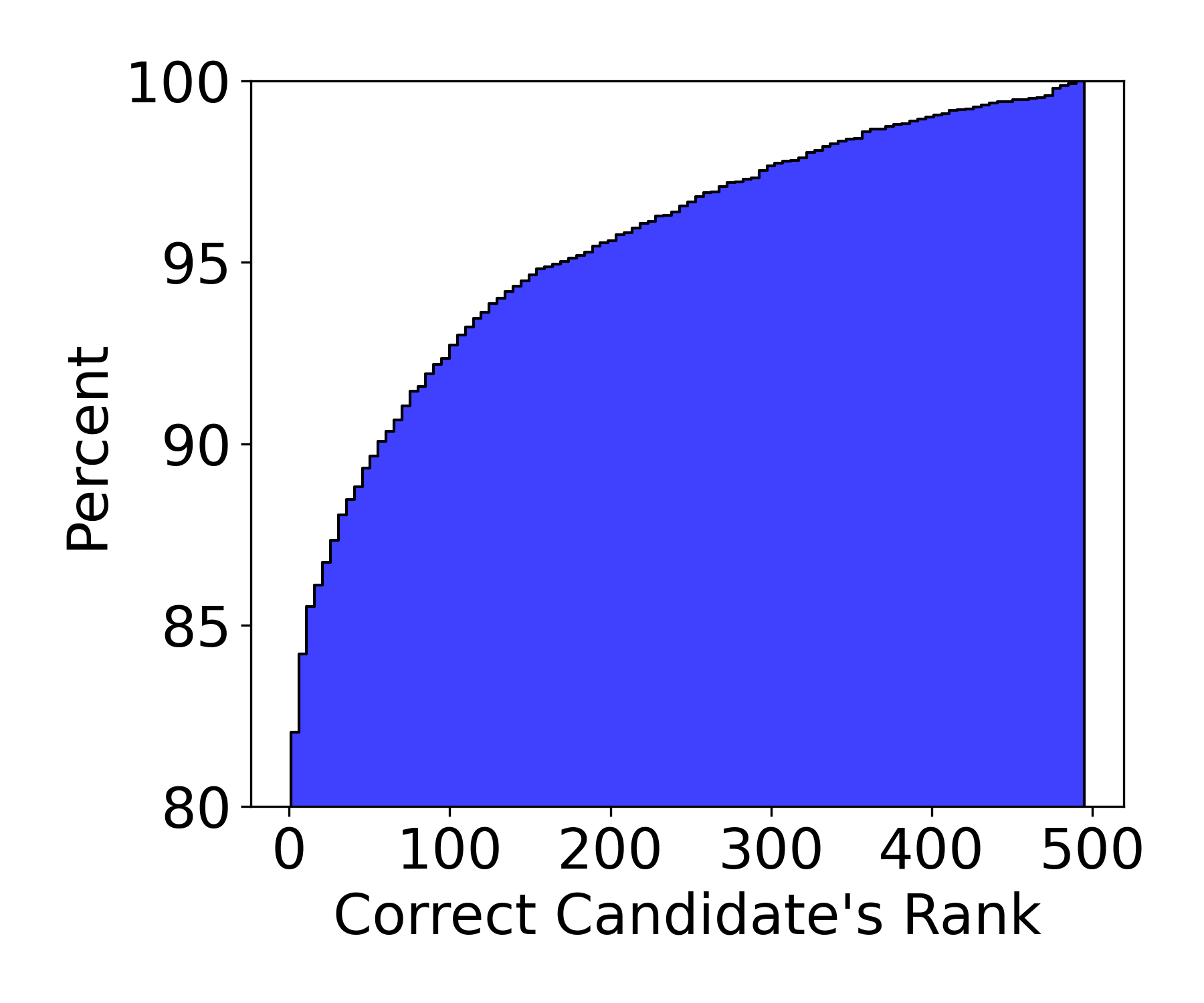}
    \caption{Correct candidate's ranking out of 500 candidates (beam search) on \textbf{exp-primitive robustness} problems.
    }
    \label{apx:fig:beam500-correctrank}
\end{figure}

We study the simple-primitive robustness problems (from \autoref{tbl:robust-simple}, top), as these problems resulted in a small number of \texttt{Sympy} timeouts, allowing for scaling of verification.
We increase the beam-width to 500 candidates, and show \texttt{Fail@k} for $k\in \{1,10,50,200,500\}$ in \autoref{apx:fig:beam500}.
The top of the solution list remains brittle with the larger beam size: that is, with beam size 500 the top-ranked solutions are still frequently incorrect, supporting the findings in \autoref{tbl:search}.
However, the correct solution often exists deeper into the ranked list, i.e. \texttt{Fail@k} decreases as $k$ increases, ranging from 0.1 \texttt{Fail@500} for \texttt{cos} to 3.7 \texttt{Fail@500} for \texttt{tan}. 
As an illustration, \autoref{apx:fig:beam500-correctrank} shows the ranking of the correct solution for $\texttt{exp}$-robustness problems that had a solution in the top-500 beam candidates.

\subsection{Alternate Decoding Strategies}

So far, we have used beam search, a deterministic search that approximately searches for the highest scoring output sequence.
An alternative approach is decoding a set of sequences $\{\hat\yb_1,\ldots,\hat\yb_k\}$ by \textit{sampling} recursively from  model-dependent per-token distributions, $y_t^{(i)}\sim q(y_t|y_{<t},\xb, p_\theta)$,
where each $\hat\yb_i=(y_1^{(i)},\ldots,y_{T_i}^{(i)})$.
We use the common approach of temperature sampling.

\begin{figure}
    \centering
    \includegraphics[width=\columnwidth]{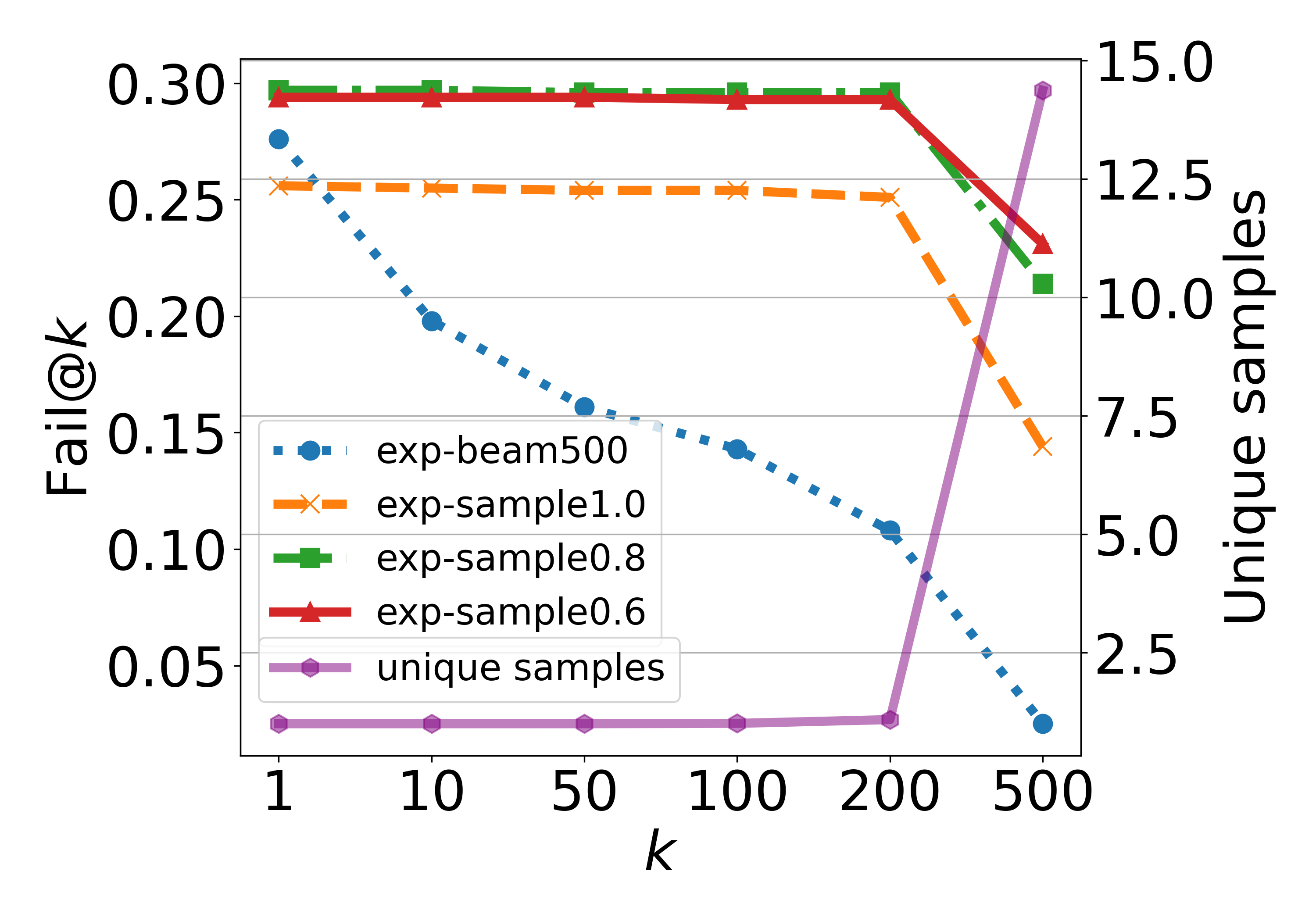}
    \caption{Comparing \texttt{Fail@k} using \textbf{sampling vs. beam search} on \textbf{exp-robustness}.
    Beam search outperforms sampling, which we attribute  to better exploration with beam search: namely, beam search returns unique candidates, while sampling returns many duplicates in this setting.
    }
    \label{apx:fig:beamsample}
\end{figure}

\autoref{apx:fig:beamsample} shows \texttt{Fail@k} rates for sampling 500 candidates at temperatures $\{0.6,0.8,1.0\}$, and beam search with 500 candidates, for  $\texttt{exp}$-robustness problems (other simple-primitives gave similar results).
Beam search outperforms sampling, which we attribute to better exploration with beam search: namely, beam search guarantees unique candidates, while sampling returns many duplicates in this setting (here, only 14 unique sequences out of 500 samples).

\subsection{SAGGA -- Quantitative Summary}

\begin{figure}
\begin{center}
\includegraphics[width=\columnwidth]{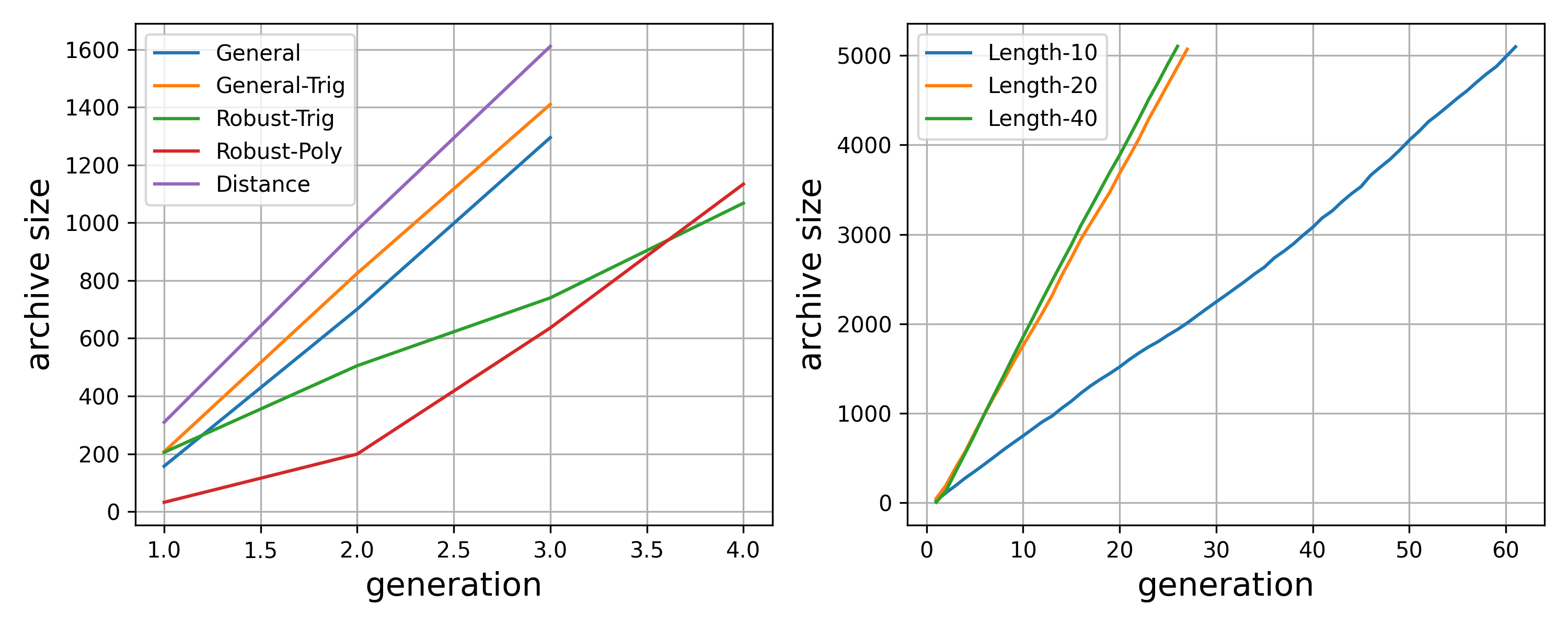}
\caption{\label{fig:genetic} Cumulative failures found per iteration of SAGGA, for various mutation and fitness settings.}
\end{center}
\end{figure}

\autoref{tbl:genetic-general} provides a quantitative summary of the archives discovered with SAGGA.
\autoref{fig:genetic} shows the rate at which SAGGA fills its archive with failure cases under different mutation strategies and fitness functions.

\begin{table}
\setlength{\tabcolsep}{1pt}
\begin{center}
\footnotesize
\begin{tabular}{lcrrrrrr}
\toprule
{} & \textbf{Iters} & \textbf{Len} & \textbf{Nodes} & \textbf{Depth} & \textbf{1-term} & \textbf{2-term} & \textbf{3-term} \\
\midrule
\textbf{General     } &              3 &         15.4 &            7.3 &            3.6 &            64.8 &            24.7 &            10.5 \\
\textbf{General-Trig} &              3 &         21.8 &           11.1 &            4.9 &            55.7 &            36.6 &             7.7 \\
\textbf{Robust-Trig } &              4 &         16.4 &            7.7 &            4.5 &            69.9 &            30.1 &             0.0 \\
\textbf{Robust-Poly } &              4 &         23.6 &           12.6 &            3.9 &            17.0 &            24.1 &            43.7 \\
\textbf{Distance    } &              3 &         30.0 &           15.0 &            5.4 &            32.6 &            34.2 &            20.2 \\
\midrule
\textbf{Length-10   } &             61 &         11.1 &            5.3 &            3.1 &            91.3 &             8.6 &             0.1 \\
\textbf{Length-20   } &             27 &         21.4 &           10.0 &            4.2 &            58.0 &            35.5 &             6.4 \\
\textbf{Length-40   } &             26 &         40.6 &           16.7 &            5.5 &            13.5 &            49.3 &            34.3 \\
\bottomrule
\end{tabular}
\caption{\label{tbl:genetic-general} Summary of the archives discovered with SAGGA for the target failure type shown on the left. 
Iters gives the number of SAGGA iterations needed to reach 1,000 failures, and 5,000 failures for the Length-X settings.
We use the number of \texttt{+} operations plus 1 as a rough proxy for the number of `terms' in a problem; e.g. $x^2+2$ has 2 terms.
}
\end{center}
\end{table}

\section{Additional Experiment Details}
\subsection{Experimental Setup}
We use the implementation and pre-trained model from \citet{lample2019deep} for all of our experiments, specifically the \texttt{FWD+BWD+IBP} model which obtained top-10 accuracies of 95.6\%, 99.5\%, and 99.6\% on their publicly available test sets.\footnote{\url{https://github.com/facebookresearch/SymbolicMathematics/}, commit \texttt{4596d07}.}
Our evaluation is based on their code provided in \texttt{beam\_integration.ipynb}. We use their utilities for inputs and outputs, and by default use beam search with beam size 10. 
When computing Fail@50 we use beam size 50.
Following \citet{lample2019deep}, we use Sympy to check whether the derivative of a prediction is equal to the original problem, \texttt{simplify(derivative - f) == 0}.
We generously count the prediction as correct if a timeout occurs.
Since Sympy's \texttt{simplify} function is imperfect, there is a possibility of false negatives, which is an inherent limitation of verifying whether a solution is correct. However, an answer that is not feasibly verifiable in a search-and-verify setting is incorrect for practical purposes.
In preliminary experiments, we found that additionally simplifying the derivative and the function before subtracting tended to reduce timeouts, and
found that Sympy's ability to simplify hyperbolic functions (e.g. $\sinh, \cosh$) was limited, so we discard problems with hyperbolic functions from our analysis.
Future work could verify with additional libraries, and we release all predictions for inspection.

\subsection{Robustness}
\paragraph{Test accuracy does not imply robustness.}
For $X_{\mathcal{N}_1}$ we sample 100 successful validation problems $f$ and $10$ values of $k$, while for $X_{\mathcal{N}_2}$ we sample 1000 successful validation problems.

\subsection{Compositionality}
\paragraph{Validation problems.} Fail@50 \texttt{valid} had abnormally many timeouts compared to the other experiments; for this setting only we do not consider a timeout as correct.

\subsection{Extrapolation}
\paragraph{More details of integer distribution.}
As a rough picture of integer values in the training distribution, we sample 100,000 sequences from each of the \texttt{FWD} and \texttt{BWD} train sets, convert them to Sympy, and count Sympy equation tree nodes of type \texttt{Integer}.
We find that 99.4\% of the positive integers were in the range [1, 100), and that other magnitude buckets were non-empty but sparsely represented -- e.g. [100, 200) had 0.2\%, and [1000, 2000) had 0.096\% ,and [10000, 20000) had 0.005\% 
of the positive integers in the problem sample.

\subsection{Runtime vs. Candidates}
We generously stop early if a successful candidate is found, and use a 1-second timeout for \texttt{Sympy}; thus the results are a lower-bound on the runtime.
In the worst case, for $N$ problems, without early stopping and with a $t$-second timeout, verification takes $O(ktN)$ seconds, e.g. 41.6 hours to verify 50 candidates on 3,000 problems with a 1 second timeout.

\section{SAGGA}

\paragraph{Default fitness.}
Unless otherwise noted, we use a fitness that favors short (and hence interpretable) problems,
\begin{align}
    F(f_\theta, \xb) &= m(\xb,f_\theta(\xb; 1))\cdot \frac{1}{|\xb|},
\end{align}
which is positive when the model returns an incorrect integral of problem $\xb$, and higher for shorter problems.

\paragraph{Mutations.}
A mutation transforms a problem's equation tree; i.e. $h(T_\xb)\rightarrow T'$ where $T_\xb$ is the equation tree of problem $\xb$.
SAGGA supports mutations for internal tree nodes and leaf nodes.

The \textit{internal} node mutations replace a node $v_i$ with:
\begin{itemize}
    \item \textbf{Constant}: an integer $k\sim \mathcal{U}(v_\text{min}, v_\text{max})$.
    \item \textbf{Symbol}: $x$.
    \item \textbf{Operation}: $v_i'\sim \{+, *, ** \}$
    \item \textbf{Add-arg}: $v_i'(w_{1:j},w_{j+1})$, where $w_{1:j}$ are the previous arguments to $v_i$ and $w_{j+1}$ is a random simple operation (see below). For instance, $\texttt{sum}(1, 2)\rightarrow \texttt{sum}(1, 2, 3x^2)$.
    If $v_i$ is a one-argument function, this mutation adds a new argument via a sum, e.g. $\texttt{exp}(1)\rightarrow \texttt{exp}(1+3x^2)$.
\end{itemize}

The \textit{leaf} node mutations replace a node $v_i$ with:
\begin{itemize}
    \item \textbf{Constant}: an integer $k\sim \mathcal{U}(v_\text{min}, v_\text{max})$.
    \item \textbf{Symbol}: $kx$ where $k\sim \mathcal{U}(v_\text{min}, v_\text{max})$.
    \item \textbf{Simple-op}: random simple operation (see below).
\end{itemize}

Each random simple operation is of the form $k_1\circ x ^ k_2$, where $k_1\sim \mathcal{U}(v_\text{min}, v_\text{max})$, $\circ\sim \{*,**,/\}$, $k_2\sim \{1, 2\}$. For instance, $3x^2$, $5/x$.

\paragraph{Default settings.}
Unless otherwise noted, we run SAGGA with the following default settings:
\begin{itemize}
    \item Beam size: 10
    \item Evaluation $m(\xb,\cdot)$: Fail@1
    \item Seed size: 100
    \item Generation size: 1000
    \item Seed selection: k-means, $k=10$
    \item Fitness threshold $\tau$: 0.01
    \item Target archive size: 1000
    \item Integer perturbation range $v_\text{min},v_\text{max}$: [-1000, 1000]
    \item Mutations: all
    \item Seed problems: $\{1, \quad x,\quad x+1,\quad x^2 + x + 1\}$
\end{itemize}
Each generation (iteration) takes around 4 minutes on a single Quadro RTX 8000 GPU.

\subsection{Robustness.}
\paragraph{Settings.}
\begin{itemize}
    \item Integer perturbation range $v_\text{min},v_\text{max}$: [-100, 100]
    \item Mutations:  only \textbf{Constant} mutations.
\end{itemize}

\paragraph{Seeds.}
Polynomial robustness:
\begin{align*}
    X_{\texttt{poly}}= \{&1, 2x, 2/x, 2x+1, 2/x+1, \\
    &2x^2+2x+1, 2x^2+2/x+1,\\
    &2x^3 + 2x^2+1, \\
    &2x^{42}+2x^3+2x^2+1\}
\end{align*}

Trigonometric robustness:
\begin{align*}
    X_{\texttt{trig}}= \{&17\cos(83x),17\cos(83x)+1,\\ &34\sin(77x),34\sin(77x)+1\\
    &2\cos(2x)+2x, 2\cos(2x)+2x+1,\\
    &2\sin(2x)+2x, 2\sin(2x)+2x+1\\
    &2\sin(2x)\cos(2x)\}
\end{align*}

In these experiments, the structure of the seeds remains fixed and the integers are varied.
The seed elements with non-trivial integers were selected based on the manual neighborhood experiments.
Note that this simply accelerates the experiment, as the algorithm could discover these.

\subsection{Out-of-distribution.}
\textbf{General failures.} We run SAGGA in two settings. The first is with default settings (see above).
The second biases the search towards trigonometric functions, using the seed,
\begin{align*}
    X_{\texttt{trig-general}}= \{&2\cos(2x),2\cos(2x)+1\\ &2\sin(2x),2\sin(2x)+1\\
    &2\cos(2x)+2x, 2\cos(2x)+2x+1,\\
    &2\sin(2x)+2x, 2\sin(2x)+2x+1\\
    &2\sin(2x)\cos(2x)\}
\end{align*}
and set the default fitness to zero unless the problem contains a trigonometric function.

\textbf{Target lengths.} We use the fitness,
\begin{align}
    F(f_\theta, \xb) &= m(\xb,f_\theta(\xb; 1))\cdot \frac{1}{|\xb - \ell|},
\end{align}
where $\ell$ is the target length.

To more clearly compare growth rates, compared to the other experiments we use a smaller seed size of 50 and smaller generation size of 300, and generate 5,000 failures rather than 1,000, resulting in more iterations per algorithm setting.

\begin{table*}
\begin{center}
\scriptsize
\setlength{\tabcolsep}{4pt}
\begin{tabular}{llll}
\toprule
                                  x &                                     hyp\_simplified &                                         derivative &                                         difference \\
\midrule
                      \textbf{-103} &                                             -103*x &                                               -103 &                                                  0 \\
                               -104 &                                        -48*x - 625 &                                                -48 &                                                 56 \\
                               -136 &                                             -128*x &                                               -128 &                                                  8 \\
                                -33 &                                        -31*x - 256 &                                                -31 &                                                  2 \\
\hline
                \textbf{2*x**(42)+21} &                               x*(2*x**42 + 903)/43 &                                       2*x**42 + 21 &                                                  0 \\
                       2*x**(42)+22 &                               2*x*(x**42 + 469)/43 &                                   2*x**42 + 938/43 &                                              -8/43 \\
                       2*x**(42)+28 &                               2*x*(x**42 + 614)/43 &                                  2*x**42 + 1228/43 &                                              24/43 \\
                       2*x**(42)+68 &                              2*x*(x**42 + 1502)/43 &                                  2*x**42 + 3004/43 &                                              80/43 \\
\hline
                \textbf{-47 + 2/x - 2/x**70} &                    -47*x + 2*log(x) + 2/(69*x**69) &                                -47 + 2/x - 2/x**70 &                                                  0 \\
                -47 + 2/x - 2/x**71 &                    -47*x + 2*log(x) + 2/(35*x**70) &                                -47 + 2/x - 4/x**71 &                                           -2/x**71 \\
               -47 + 2/x - 31/x**71 &                       -47*x + log(x**2) + x**(-60) &                               -47 + 2/x - 60/x**61 &                              (31 - 60*x**10)/x**71 \\
               -71 + 36/x - 2/x**71 &                   -71*x + 36*log(x) + 1/(17*x**70) &                         -71 + 36/x - 70/(17*x**71) &                                     -36/(17*x**71) \\
\hline
                       \textbf{13*cos(18*x)} &                                    13*sin(18*x)/18 &                                       13*cos(18*x) &                                                  0 \\
                       13*cos(19*x) &                                          sin(19*x) &                                       19*cos(19*x) &                                        6*cos(19*x) \\
                       13*cos(83*x) &                                       sin(83*x)/17 &                                    83*cos(83*x)/17 &                                  -138*cos(83*x)/17 \\
                       17*cos(47*x) &                                          sin(47*x) &                                       47*cos(47*x) &                                       30*cos(47*x) \\
\hline
                  \textbf{13*cos(82*x) - 59} &                            -59*x + 13*sin(82*x)/82 &                                  13*cos(82*x) - 59 &                                                  0 \\
                  13*cos(83*x) - 59 &                               -59*x + sin(83*x)/13 &                               83*cos(83*x)/13 - 59 &                                   -86*cos(83*x)/13 \\
                  17*cos(37*x) - 49 &                                  -49*x + sin(37*x) &                                  37*cos(37*x) - 49 &                                       20*cos(37*x) \\
                  17*cos(41*x) - 45 &                                  -45*x + sin(41*x) &                                  41*cos(41*x) - 45 &                                       24*cos(41*x) \\
\hline
              \textbf{10*sin(45*x)*cos(2*x)} &                   -5*cos(43*x)/43 - 5*cos(47*x)/47 &                          5*sin(43*x) + 5*sin(47*x) &                                                  0 \\
              10*sin(47*x)*cos(2*x) &           -255*cos(45*x)/2201 - 215*cos(49*x)/2201 &        11475*sin(45*x)/2201 + 10535*sin(49*x)/2201 &            470*sin(45*x)/2201 - 470*si... \\
              10*sin(90*x)*cos(2*x) &                   -5*cos(43*x)/44 - 5*cos(47*x)/46 &                215*sin(43*x)/44 + 235*sin(47*x)/46 &  215*sin(43*x)/44 + 235*sin(47... \\
              19*sin(90*x)*cos(2*x) &                            cos(2*x)**2*cos(90*x)/2 &  -2*sin(2*x)*cos(2*x)*cos(90*x) - 45*sin(90*x)*... &  -(43*sin(88*x)/2 + 19*sin... \\
\hline
          \textbf{-x**2 + x + log(4)*tan(x)} &              -x**3/3 + x**2/2 - log(4)*log(cos(x)) &                   -x**2 + x + log(4)*sin(x)/cos(x) &                                                  0 \\
    -x**2 + x + log(4)*tan(17*x**2) &  -x**3/3 + x**2/2 + log(2)*log(cos(17*x**2)... &   -x**2 + 2*x*log(2)*sin(17*x**2)/cos(17*x**2) + x &                        (x - 1)*log(4)*tan(17*x**2) \\
     -x**2 + x + log(4)*tan(2*x**2) &  -x**3/3 + x**2/2 + log(2)*log(cos(2*x**2)... &     -x**2 + 4*x*log(2)*sin(2*x**2)/cos(2*x**2) + x &                       (2*x - 1)*log(4)*tan(2*x**2) \\
    -x**2 + x + log(4)*tan(63/x**2) &  -x**3/3 + x**2/2 - log(2)*log(cos(63/x**2)... &  -x**2 + x + 252*log(2)*sin(63/x**2)/(x**3*cos(... &              (126 - x**3)*log(4)*tan(63/... \\
\hline
                  \textbf{sqrt(3*x + 3) - 2} &                  -2*x + 2*sqrt(3)*(x + 1)**(3/2)/3 &                            sqrt(3)*sqrt(x + 1) - 2 &                                                  0 \\
      sqrt(-86**(x**2) + 62/x) - 40 &                     -40*x + acosh(86**(-x**2/2)/x) &  -40 + (-86**(-x**2/2)*log(86) - 86**(-x**2/2)/... &  -sqrt(-86**(x**2) + 62/x) ... \\
                sqrt(14 + 62/x) + 4 &  (49*x**(3/2) + 217*sqrt(x) + sqrt(7*x + 31... &  (147*sqrt(x)/2 + (28 + sqrt(749)/(sqrt(x)*sqrt... &  (-x*(7*x + 31)**(5/2)*sqrt(2... \\
                sqrt(14 + 62/x) - 2 &  (49*x**(3/2) + 217*sqrt(x) + (-14*x + 31*... &  (147*sqrt(x)/2 + (-14 + sqrt(854)/(2*sqrt(x)*s... &  (x*(7*x + 31)**(5/2)*sqrt(6... \\
\hline
                 \textbf{tan(exp(2))/(18*x)} &                              log(x)*tan(exp(2))/18 &                                 tan(exp(2))/(18*x) &                                                  0 \\
  tan(26*x + 2 + exp(2))/(18*x**75) &  [add, mul, div, INT-, 1, INT+, 1, 5, 1, 2, mul... &                                                 -- &                                                 -- \\
        tan(exp(-9504*x**2))/(18*x) &             -log(cos(exp(-9504*x**2))**(-2))/18144 &  44*x*exp(-9504*x**2)*sin(exp(-9504*x**2))/... &  44*x*exp(-9504*x**2)*sin(... \\
             tan(exp(-96*x))/(18*x) &                   -log(cos(exp(-96*x))**(-2))/1728 &     exp(-96*x)*sin(exp(-96*x))/(9*cos(exp(-96*x))) &  (2*x - exp(96*x))*exp(-9... \\
\bottomrule
\end{tabular}
\caption{\label{apx:tbl:sagga-robustness}\textit{SAGGA robustness}. We show the problem $\xb$, the simplified prediction from the neural sequence integrator, its derivative, and the derivative's difference with $\xb$. The prediction is incorrect when the difference is not zero.
When the model's prediction does not parse successfully, we show its unparsed infix tokens.
We show a nearby problem that the model gets correct in \textbf{bold} (either hand-selected or a validation example); all other problems are failures discovered by SAGGA.
}
\end{center}
\end{table*}

\begin{table*}
\begin{center}
\scriptsize
\begin{tabular}{llp{5cm}p{4cm}p{4cm}}
\toprule
{} &                     x &                                 hyp\_simplified &                                                                                                                                                    derivative &                                                                   difference \\
\midrule
0  &                 \textbf{30**x} &                                  30**x/log(30) &                                                                                                                                                         30**x &                                                                            0 \\
1  &                119**x &                                   119**(x - 1) &                                                                                                                                         119**(x - 1)*log(119) &                                               119**(x - 1)*(-119 + log(119)) \\
2  &                132**x &                          132**x/(1 + log(132)) &                                                                                                                                132**x*log(132)/(1 + log(132)) &                                                       -132**x/(1 + log(132)) \\
3  &                136**x &                                exp(x*log(136)) &                                                                                                                                      exp(x*log(136))*log(136) &                                           -136**x + exp(x*log(136))*log(136) \\
\hline
4  &             -100*x**x &                                       -50*x**2 &                                                                                                                                                        -100*x &                                                            -100*x + 100*x**x \\
5  &             -149*x**x &                                    -149*x**2/2 &                                                                                                                                                        -149*x &                                                            -149*x + 149*x**x \\
6  &             -151*x**x &                                    -151*x**2/2 &                                                                                                                                                        -151*x &                                                            -151*x + 151*x**x \\
\hline
7  &   158*x**(x**2) + 611 &                158*x**2/sqrt(x**2 + 1) + 611*x &                                                                                                      -158*x**3/(x**2 + 1)**(3/2) + 316*x/sqrt(x**2 + 1) + 611 &           -158*x**3/(x**2 + 1)**(3/2) + 316*x/sqrt(x**2 + 1) - 158*x**(x**2) \\
8  &   256*x**(x**2) + 191 &  x*(191*x**2 + 256*x**(x**2) + 191)/(x**2 + 1) &  -2*x**2*(191*x**2 + 256*x**(x**2) + 191)/(x**2 + 1)**2 + x*(382*x + 256*x**(x**2)*(2*x*log(x) + x))/(x**2 + 1) + (191*x**2 + 256*x**(x**2) + 191)/(x**2 + 1) &             512*x**(x**2 + 2)*(x**2*log(x) + log(x) - 1)/(x**4 + 2*x**2 + 1) \\
9  &   332*x**(x**2) + 559 &          x*(559*x**2 + 332*x + 559)/(x**2 + 1) &                                            -2*x**2*(559*x**2 + 332*x + 559)/(x**2 + 1)**2 + x*(1118*x + 332)/(x**2 + 1) + (559*x**2 + 332*x + 559)/(x**2 + 1) &  332*(2*x - x**(x**2) - 2*x**(x**2 + 2) - x**(x**2 + 4))/(x**4 + 2*x**2 + 1) \\
\hline
10 &  -240**x + 2*cos(2*x) &                    -exp(x*log(240)) + sin(2*x) &                                                                                                                        -exp(x*log(240))*log(240) + 2*cos(2*x) &                                            240**x - exp(x*log(240))*log(240) \\
11 &  -398**x + 2*cos(2*x) &                    -exp(x*log(398)) + sin(2*x) &                                                                                                                        -exp(x*log(398))*log(398) + 2*cos(2*x) &                                            398**x - exp(x*log(398))*log(398) \\
12 &  -692**x + 2*sin(2*x) &                    -exp(x*log(692)) - cos(2*x) &                                                                                                                        -exp(x*log(692))*log(692) + 2*sin(2*x) &                                            692**x - exp(x*log(692))*log(692) \\
\bottomrule
\end{tabular}
\caption{\label{apx:tbl:sagga-exponent}\textit{SAGGA OOD -- $g(x)^x$}. We show the problem $\xb$ discovered by SAGGA, the simplified prediction from the neural sequence integrator, its derivative, and the derivative's difference with $\xb$. The prediction is incorrect when the difference is not zero.
For the first cluster, we manually found a nearby problem that the model gets correct in \textbf{bold}; thus this cluster can be seen as a \textit{robustness} failure.
The second two clusters involve $x^x$ or $x^{x^2}$ which to our knowledge do not have analytical integrals (e.g. see \url{https://www.wolframalpha.com/input/?i=integral+of+x**x}); these clusters can be seen as \textit{exploits}.
}
\end{center}
\end{table*}

\begin{table*}
\begin{center}
\scriptsize
\begin{tabular}{llp{7cm}p{7cm}l}
\toprule
{} &                              x & hyp\_raw & Correct Integral (\texttt{Wolfram Alpha})\\
\midrule
0 &                 169*sin(4*x)/x &  169*log(x**2)/8 - 169*cos(4*x)/4 & $169\ \mathrm{Si}(4 x)$\\
1 &                   -2*sin(42/x) &  cos(42/x)/21 & $-2 (-42 \mathrm{Ci}(42/x) + x \sin(42/x))$\\
2 &      -2*sin(185*x**2)*cos(2*x) &  4*sin(2*x)*sin(185*x**2)/3421 + 370*cos(2*x)*cos(185*x**2)/3421 & \texttt{-(Sqrt[Pi/370] (Cos[1/185] FresnelS[Sqrt[2/(185 Pi)] (-1 + 185 x)] + Cos[1/185] FresnelS[Sqrt[2/(185 Pi)] (1 + 185 x)] - (FresnelC[Sqrt[2/(185 Pi)] (-1 + 185 x)] + FresnelC[Sqrt[2/(185 Pi)] (1 + 185 x)]) Sin[1/185]))}\\
3 &  357**(x**(2**x)) + 2*sin(2*x) &  exp(x*log(2)) - cos(2*x) & No result \\
4 &          1/(x**48*(3*x+2)**49) &  add, mul, div, INT-, 1, 0, 9, 0, 9, 1, 9, 9, 9, 9, 9, 9, 9, 9, 9, 9, 9, 9, 9, 5, INT+, 1, 0, 2, 4, ln, x, add, mul, div, INT+, 1, 0, 9, 0, 9, 1, 9, 9, 9, 9, 9, 9, 9, 9, 9, 9, 9, 9, 9, 9, 9, 9, 9, 9, 9, INT+, 1, 0, 2, 4, ln, add, div, INT+, 2, INT+, 3, x, mul, INT-, 1, mul, pow, add, mul, INT+, 1, 0, 0, 9, 6, 2, 9, 4, 4, 0, 0, 0, 0, 0, 0, 0, 0, 0, 0, 0, 0, 0, 0, 0, 0, 0, 0, 0, 0, 0, 0, 0, 0, 0, 0, 0, 0, 0, 0, 0, 0, 0, 0, 0, 0, 0, 0, 0, 0, 0, 0, 0, 0, 0, 0, 0, 0, 0, 0, 0, 0, 0, 0, 0, 0, 0, 0, 0, 0, 0, 0, 0, 0, 0, 0, 0, 0, 0, 0, 0, 0, 0, 0, 0, 0, 0, 0, 0, 0, 0, 0, 0, 0, 0, 0, 0, 0, 0, 0, 0, 0, 0, 0, 0, 0, 0, 0, 0, 0, 0, 0, 0, 0, 0, 0, 0, 0 & See https://www.wolframalpha.com/input/? i=integral+of+1\%2F\%28x**48*\%283*x\%2B2\%29**49\%29\\
\bottomrule
\end{tabular}
\caption{\label{apx:tbl:sagga-exploits}\textit{SAGGA OOD -- exploits}. We show the problem $\xb$, and the neural sequence integrator's prediction as either raw prefix tokens or (when possible) in its infix form. 
All problems are failures discovered by SAGGA.
}
\end{center}
\end{table*}

\end{document}